\newif\ifreview
\reviewfalse 

\newif\ifarxiv
\arxivtrue 

\newif\ifsubmission
\ifreview\submissiontrue\else\submissionfalse\fi

\documentclass[10pt,twocolumn,letterpaper]{article}

\ifreview
  \usepackage[review]{cvpr}
\else
  \ifarxiv
    \usepackage[pagenumbers]{cvpr}
  \else
    \usepackage{cvpr}
  \fi
\fi

\ifsubmission
\PassOptionsToPackage{hypertexnames=false}{hyperref}
\fi
\definecolor{cvprblue}{rgb}{0.21,0.49,0.74}
\usepackage[pagebackref,breaklinks,colorlinks,allcolors=cvprblue]{hyperref}

\usepackage[capitalize]{cleveref}
\usepackage{caption}
\usepackage{subcaption}
\usepackage{graphicx}
\usepackage{hyperref}

\usepackage{lipsum}
\usepackage{wrapfig}
\usepackage{array}
\usepackage{xcolor}
\usepackage{makecell}
\usepackage{enumitem}

\usepackage{algorithm}
\usepackage{algpseudocode}
\floatname{algorithm}{Algorithm}
\algrenewcommand\algorithmicrequire{\textbf{Require:}}
\algrenewcommand\algorithmicensure

\usepackage{colortbl}
\usepackage{multirow}
\usepackage{multicol}
\usepackage{listings} 
\usepackage{titletoc} 
\usepackage{fancyvrb}
\usepackage{tcolorbox} 
\usepackage[accsupp]{axessibility}  
\usepackage[subtle]{savetrees} 

\makeatletter
\newcommand{\subpara}[1]{%
\if@nobreak\else\vspace{1mm}\fi
\noindent\textbf{#1}%
}
\makeatother

\definecolor{lightgray}{rgb}{0.83, 0.83, 0.83}
\definecolor{Gray}{gray}{0.6}
\definecolor{aliceblue}{rgb}{0.94, 0.97, 1.0}
\definecolor{mistyrose}{rgb}{1.0, 0.89, 0.88}
\definecolor{backcolour}{rgb}{0.95,0.95,0.92}

\crefname{equation}{Eq.}{Eqs.}
\Crefname{equation}{Equation}{Equations}

\crefname{figure}{Fig.}{Figs.}
\Crefname{figure}{Figure}{Figures}

\crefname{table}{Tab.}{Tabs.}
\Crefname{table}{Table}{Tables}

\crefname{section}{Sec.}{Secs.}
\Crefname{section}{Section}{Sections}

\crefname{algorithm}{Alg.}{Algs.}
\Crefname{algorithm}{Algorithm}{Algorithms}

\AtBeginDocument{%
  \setlength{\abovedisplayskip}{3pt}%
  \setlength{\belowdisplayskip}{3pt}%
  \setlength{\abovedisplayshortskip}{2pt}%
  \setlength{\belowdisplayshortskip}{2pt}%
}

\newcommand{\figcapabove}{\vspace{-20pt}} 
\newcommand{\figcapbelow}{\vspace{-10pt}} 
\newcommand{\tabcapbelow}{\vspace{-6pt}}  
\newcommand{\tabbelow}{\vspace{-6pt}}     

\ifsubmission
\usepackage{chapterbib}
\fi

\def\confName{CVPR}
\def\confYear{2027}

\title{VSpark: Speculative Decoding for Autoregressive Vision-Language-Action Models}

\author{
Suho Yoo$^{1\:\ast}$ \quad
Youngjoon Jang$^{1, 2}$\thanks{Equal contribution.} \quad
Hyebin Cho$^1$ \quad
Taeyeop Lee$^1$ \quad
Jaewook Jeong$^1$ \quad
Joon Son Chung$^1$ \\[0.3em]
$^1$KAIST \quad$^2$VGG, University of Oxford
\\[0.3em] \href{https://suho-yoo.github.io/VSpark/}{\textcolor[HTML]{1F4E8C}{https://suho-yoo.github.io/VSpark/}}
}
\makeatletter
\g@addto@macro\@maketitle{%
  \begin{minipage}{\textwidth}
  \captionsetup{type=figure}
  \centering
  \includegraphics[width=\linewidth]{figures/main_teaser.pdf}
  \vspace{-15pt}
    \caption{\textbf{Accelerating VLAs with VSpark.}
    (a) VSpark accelerates autoregressive vision-language-action models by up to $2.10\times$ on LIBERO.
    (b) Different action token sequences can represent similar robot motions, so predictions along a rejected branch often still hold under the corrected prefix.
    Conventional speculative decoding discards them, whereas VSpark reuses them to improve drafting while retaining exact target verification.}
    \label{fig:teaser}
    \end{minipage}
    \par\vspace{30pt}
}
\makeatother

\makeatletter
\newcommand{\savesuppcitationunit}{\ifsubmission\global\let\supp@citationunit\the@ipfilectr\fi}
\newcommand{\restoresuppcitationunit}{\ifsubmission\global\let\the@ipfilectr\supp@citationunit\fi}
\makeatother
\begin{document}
\ifsubmission
  
\maketitle
\begin{abstract}
Our aim is to accelerate autoregressive vision-language-action models (VLAs), which generate robot actions one token at a time. Speculative decoding makes this possible: a lightweight draft model proposes several tokens at once, and the VLA verifies them in a single pass, accepting only those that match its own predictions, so decoding remains lossless. The speedup, however, is limited in robot control, where many different tokens encode nearly the same motion and a draft close to the correct action is still rejected. To address this, our approach, VSpark, is built upon three design choices: (i) rollouts collected with calibrated action noise, which expose the draft to a wider range of robot states; (ii) feature alignment, which trains the draft to match the VLA's hidden features rather than its output tokens alone; and (iii) continuation reuse, which keeps the predictions the VLA has already computed along rejected drafts as candidates for the next verification step, instead of discarding them. On four VLAs with different action tokenizers and four manipulation benchmarks, VSpark achieves 1.28$\times$ to 2.10$\times$ speedup while keeping the VLA's decoding lossless.
\end{abstract}

\section{Introduction}
\label{sec:intro}

Vision-language-action models (VLAs) control robots by mapping camera images and a language instruction to actions, building on pretrained vision-language models~\cite{rt-2, openvla}. Autoregressive VLAs represent continuous actions as discrete tokens and generate them by the next-token prediction used in pretraining, which lets them inherit the generalization and visual-language understanding of the underlying model~\cite{chen2023pali, driess2023palm, karamcheti2024prismatic}. The cost is speed: every action token requires a forward pass of a large model, limiting the control frequency.

Progress in action tokenization has reduced this cost but not removed it. Early models assign one token to every action dimension at every timestep~\cite{openvla}, whereas recent action tokenizers compress a whole chunk of future actions into a shorter token sequence, so that a single token no longer corresponds to a single action value~\cite{fast,actioncodec,g05}. The remaining tokens are still generated one at a time, and this remains a major bottleneck.

The focus of this work is on removing this bottleneck with speculative decoding~\cite{leviathan2023,chen2023}. A lightweight draft model proposes several future tokens, and the VLA being accelerated, the \emph{target}, verifies them in one parallel pass and accepts the longest prefix that matches its own predictions. Because only matching tokens are accepted, decoding is lossless through exact verification by the target. This property is valuable in robot control: the accelerated policy uses lossless decoding, with no tolerance to tune and exact verification of action tokens. The speedup, however, depends on how many drafted tokens are accepted per verification cycle, which is low for robot actions. Many different tokens encode nearly the same motion, so a draft close to the correct action is still rejected when its tokens differ.


Existing speculative decoding methods for VLAs relax verification by accepting drafted tokens that are sufficiently close to the target's predictions~\cite{specvla,kerv,heisd}. These acceptance rules rely on per-dimension action bins, where neighboring bins correspond to nearby values of the same action dimension~\cite{rt-2,openvla}. This notion of proximity does not directly extend to compressed tokenizers such as FAST~\cite{fast} and ActionCodec~\cite{actioncodec}, which require the complete token sequence before decoding it into an action chunk. Direct comparison with the target's decoded actions would require the target to generate the remaining tokens of the action chunk before acceptance can be decided. In other words, relaxing acceptance for a single token would require the target to generate the entire remaining action sequence.

Token embedding distance offers an alternative that can be evaluated immediately, but we find that it is an unreliable proxy for action distance. Ranking candidate tokens by embedding distance disagrees with ranking them by decoded action distance for about 31\% of pairs, and 18\% of single-token substitutions in complete FAST sequences do not decode to a valid action at all (\cref{fig:action_distance}). When verification is relaxed by embedding distance on $\pi_0$-FAST, raising the speedup from $1.60\times$ to $1.90\times$ lowers the success rate from 91.7\% to 71.4\% (\cref{tab:embedding_relaxed}). This raises our question: can speculative decoding be effective for VLAs with compressed action tokenizers while keeping verification exact?

To this end, we introduce VSpark, which keeps verification exact and instead exploits the similarity between actions on the draft side (\cref{fig:teaser}). If many tokens encode nearly the same motion, then the draft should learn from more than token labels, and a rejected token is often nearly right, so what the target predicted after it remains informative. VSpark builds on the parallel drafter DSpark~\cite{dspark} and rests on three design choices.

First, we train the draft on rollouts collected with calibrated action noise. Drafts are usually trained only on the states visited by the training demonstrations. 
We adapt noise injection in DART~\cite{dart} to draft training by perturbing actions during target rollouts in simulation. DART calibrates noise to discrepancies between a learned policy's actions and those of a supervisor, such as a human expert. We instead use discrepancies between draft and target actions, setting the noise scale to reflect the draft's errors in predicting the target.
Second, we introduce feature alignment. Supervision by token labels treats every wrong token alike, however close its action is to the correct one. We therefore feed the target's hidden features to the draft and train it to regress them, giving it a continuous signal that token identity alone does not provide. This raises the average number of tokens committed per verification cycle from 3.22 to 3.87 on $\pi_0$-FAST (\cref{tab:component_ablation}).
Third, we propose continuation reuse. When the target verifies a tree of candidates, it also computes what it would predict after each rejected candidate, and standard decoding discards these predictions. 
We consider draft tokens that match the target's predictions after a rejected token:
In VLAs, 45\% to 55\% of these tokens also match the target's predictions under the corrected prefix (\cref{tab:continuation_survival}).
Continuation reuse promotes these predictions into the next draft tree, where they are verified again,
ensuring lossless decoding at no extra cost to the target. Acting purely at inference time, continuation reuse requires no training and no change to either model.

The outcome of these design choices is that VSpark accelerates four autoregressive VLAs, $\pi_0$-FAST, StarVLA-FAST, ActionCodec, and G0.5, which together cover frequency-based and learned vector-quantized action tokenizers. Across LIBERO, LIBERO-Plus, LIBERO-Pro, and SimplerEnv, it achieves $1.28\times$ to $2.10\times$ speedup over autoregressive decoding while retaining exact verification of action tokens. Under the same exact verification, VSpark outperforms five recent speculative decoding methods~\cite{eagle3,peagle,dflash,dflash2,dspark} on every model and benchmark. This shows that similarity between robot actions can be turned into speed while retaining exact verification.

\section{Related Works}

\subpara{Action tokenization}
converts continuous robot actions into discrete tokens that an autoregressive VLA can predict, and maps the generated tokens back into actions. Tokenizers fall into two categories. Per-dimension tokenizers discretize each action dimension into bins: OpenVLA~\cite{openvla} predicts seven tokens per control step, three for translation, three for rotation, and one for the gripper. Compressed tokenizers instead encode a chunk of actions jointly across dimensions and timesteps, which shortens the sequence. FAST~\cite{fast} applies a discrete cosine transform to the chunk, quantizes the coefficients, and compresses them with byte-pair encoding. ActionCodec~\cite{actioncodec} and G0.5~\cite{g05} learn a discrete representation with a neural encoder and vector quantization, using residual quantization to refine the reconstruction. With a compressed tokenizer, a token carries information from several dimensions and timesteps, and the action is recovered only after the whole sequence is decoded. We evaluate VSpark on VLAs with both kinds of compressed tokenizers.

\subpara{Speculative decoding}
accelerates autoregressive generation with lossless decoding by verifying multiple draft tokens in one target pass~\cite{leviathan2023,chen2023}. Tree verification checks several alternative continuations in the same pass~\cite{specinfer,medusa}.
Developed for large language models, this line of work has concentrated on the draft model, which falls into two categories. Autoregressive drafters run once for every drafted token: EAGLE~\cite{eagle} drafts with a single Transformer block that predicts the target's hidden features, EAGLE-2~\cite{eagle2} builds the draft tree dynamically from cumulative draft scores, and EAGLE-3~\cite{eagle3} removes feature regression and trains on draft-generated contexts with multi-layer target features. Parallel drafters instead predict all positions in one pass: P-EAGLE~\cite{peagle} and PARD~\cite{pard,pard2} replace the unavailable preceding tokens with learned or mask embeddings, and DFlash~\cite{dflash} uses bidirectional attention within a masked block, injecting target features as keys and values in every draft layer. Because such predictions do not depend on the tokens chosen at earlier positions, DSpark~\cite{dspark}, DFlash2~\cite{dflash2}, and Domino~\cite{domino} add lightweight sequential corrections conditioned on the preceding tokens. Recent work also constructs draft trees from parallel drafters by selecting high-probability branches from the per-position distributions~\cite{ddtree,dominotree,cadtree,chains2trees,zhou2026treespark}. 
Other work reuses target predictions or hidden states that would otherwise be discarded after verification to improve subsequent drafts in large language models~\cite{luo2025turning,koo2026carryover}.
Speculative decoding has also been extended to vision-language models that generate text~\cite{glance,unilaspec,mmsdsurvey}. 
VSpark builds on DSpark and differs in three respects: it injects calibrated action noise during rollouts, restores the feature supervision that EAGLE-3 removed, and fills the draft tree with target-predicted continuations rather than draft predictions alone.

\subpara{Speculative decoding for action policies}
applies this procedure to the action tokens generated by a VLA. Existing methods use the tree decoding of EAGLE-2 and relax the verification rule, accepting a drafted token that differs from the target's prediction when the two are sufficiently close. Spec-VLA~\cite{specvla} accepts a drafted token when its action bin lies within a fixed distance of the target's. KERV~\cite{kerv} adjusts this tolerance with kinematic information and uses a Kalman filter to predict the remaining action dimensions after a rejection. HeiSD~\cite{heisd} switches between drafts retrieved from the training data and drafts generated by a draft model according to the recent robot motion, and accelerates retrieved drafts by skipping verification and relaxing acceptance over groups of action tokens. WA-SpecDec~\cite{waspecdec} injects world-model features into a fine-tuned target to make relaxed acceptance more reliable, and observes that the deviations it admits are most harmful near contact. Verification has also been used to decide how much of an action chunk to execute~\cite{a3commit}, and speculative inference has been explored for diffusion-based VLAs~\cite{realtimeflash,deltoris}, which generate continuous actions and fall outside token-level verification. 
The token-level methods above are all developed on the per-dimension binning of OpenVLA and relax or skip verification. 
Comparing decoded actions with compressed tokenizers requires autoregressively generating the complete target action chunk, negating speculative gains. VSpark instead maintains strict verification for lossless acceleration.

\begin{figure*}[t]
    \centering
    \includegraphics[width=\textwidth]{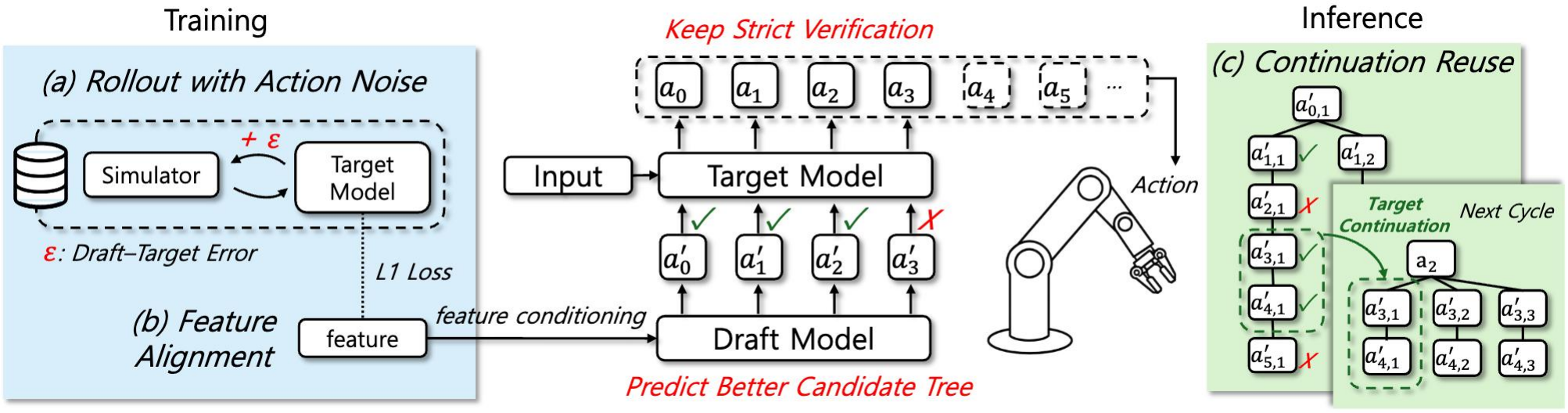}
    \vspace{-2em}
    \captionof{figure}{\textbf{Overview of VSpark.} During training, (a) rollouts with calibrated action noise collect target supervision at perturbed states, and (b) feature alignment conditions and supervises the draft with target representations. During inference, (c) continuation reuse guides the next draft tree with target continuations along rejected branches. Strict verification checks all candidates under their current prefixes, preserving lossless decoding.}
    \label{fig:main}
    \vspace{-15pt}
\end{figure*}

\section{Method}
\label{sec:method}

Given an observation and a language instruction, VSpark uses lossless decoding for the target VLA's action tokens, using fewer forward passes of the target. As shown in \cref{fig:main}, it changes how the draft is trained (\cref{sec:dart,sec:feature_draft}) and which candidates are verified in each cycle (\cref{sec:tree_fusion}), leaving the target, the action tokenizer, and the verification rule untouched.

\subsection{Preliminaries}
\label{sec:preliminaries}

\subpara{Autoregressive action generation.}
A VLA predicts actions from an observation $o$ (camera images and, when available, proprioceptive states) and a language instruction $\ell$. An autoregressive VLA generates action tokens $x_{1:L}$ one at a time,
\begin{equation}
p_{\mathcal{M}_t}(x_{1:L}\mid o,\ell)
= \prod_{i=1}^{L} p_{\mathcal{M}_t}(x_i\mid o,\ell,x_{<i}),
\label{eq:action_generation}
\end{equation}
and an action tokenizer decodes the sequence into continuous control commands. We call this model the target $\mathcal{M}_t$, and the lightweight model that drafts tokens for it the draft $\mathcal{M}_d$.

\subpara{Speculative decoding with draft trees.}
Speculative decoding replaces the $L$ sequential passes of the target by a smaller number of verification cycles. In each cycle, the draft proposes a tree of candidate tokens that extends the tokens confirmed so far, the confirmed prefix. Each node holds one candidate token, and the path from the root to a node is the context in which that token is proposed. Candidates are expanded according to the draft probabilities, and the nodes with the highest cumulative log probabilities are kept within a fixed node budget. The target then verifies all nodes in a single forward pass, using a tree attention mask that restricts each node to the confirmed prefix, its ancestors, and itself. Under greedy decoding, verification walks down the tree from the root, following at each step the child that equals the target's prediction. When no child matches, the target's prediction is appended as a correction token; when all proposals on the path match, one additional token predicted by the target is appended. The accepted tokens extend the confirmed prefix, which is the context for the next cycle. In standard tree decoding, all branches outside the accepted path are discarded, together with the predictions that the target computed for them.

\subpara{DSpark.}
We build on DSpark~\cite{dspark}, a parallel drafter. Its backbone takes the last confirmed token followed by mask embeddings, attends to projected target features of the confirmed prefix as keys and values, and produces base logits $U_i$ for all proposal positions in one bidirectional pass. As these logits ignore the tokens chosen at earlier positions, a lightweight Markov head adjusts them using the preceding token:
\begin{equation}
p_{\mathcal{M}_d}(\cdot\mid o,\ell,x_{<i})
= \operatorname{softmax}\!\left(U_i + W_1[x_{i-1}]W_2\right),
\label{eq:markov}
\end{equation}
where $W_1$ embeds the preceding token and $W_2$ projects it to vocabulary logits. This introduces a dependency between consecutive tokens without rerunning the backbone, and in a draft tree each parent token determines the adjustment applied to its children. With the target frozen, DSpark trains the draft with a cross-entropy (CE) term on the target's token, $\ell^{\mathrm{CE}}_i=-\log p_{\mathcal{M}_d}(x_i)$, and a total variation (TV) term between the draft and target distributions after softmax, $\ell^{\mathrm{TV}}_i=\lVert p_{\mathcal{M}_d}(\cdot)-p_{\mathcal{M}_t}(\cdot)\rVert_1$:
\begin{equation}
\mathcal{L}_{\mathrm{pred}}
= \sum_i w_i\left(\lambda\,\ell^{\mathrm{CE}}_i + (1-\lambda)\,\ell^{\mathrm{TV}}_i\right),
\label{eq:dspark_losses}
\end{equation}
where $w_i$ weights earlier positions more, $\lambda$ balances the two terms, and the context $(o,\ell,x_{<i})$ is omitted for brevity.

\subsection{Rollouts with Calibrated Action Noise}
\label{sec:dart}

The purpose of this step is to supervise the draft on more states than the training demonstrations contain. Following the noise injection of DART~\cite{dart}, we roll out the target in simulation while perturbing the actions it executes, so that the robot visits states near the demonstrated trajectories. DART calibrates noise to a learned policy's errors in imitating its supervisor, such as a human expert. We instead calibrate the perturbations to discrepancies between draft and target actions, reflecting the draft's objective of matching the target in speculative decoding.

\subpara{Calibrating the noise.}
At each training observation, we replace a contiguous block of the target's tokens with the draft's proposals before verification, keeping the remaining tokens fixed. We decode both sequences and measure their difference in translation and rotation, $r_n\in\mathbb{R}^{6}$. The noise covariance is the second moment of these differences,
\begin{equation}
\Sigma = \frac{1}{N}\sum_{n=1}^{N} r_n r_n^\top,
\label{eq:noise_covariance}
\end{equation}
where $N$ is the number of valid differences. Zero-mean Gaussian noise with this covariance captures both the magnitude of the draft's errors and their structure across action dimensions.

\subpara{Collecting supervision.}
Rollouts start from simulator states in the training data. At each observation, the frozen target generates action tokens and hidden features. We decode the tokens, add the calibrated noise to the translation and rotation commands, clip them to the valid action range, and leave the gripper command unchanged. Executing the perturbed action yields the next observation, where the target is queried again. Each visited observation is stored with the tokens and features that the target produced before the perturbation. The supervision is therefore clean, and the noise is used only during data collection.

\subsection{Feature Alignment}
\label{sec:feature_draft}

Feature alignment gives the draft information that token labels cannot. Under a token-level loss, a token decoding to almost the same motion as the target's is as wrong as any other. We therefore use the target's hidden features as both draft input and regression target, training on the rollouts of \cref{sec:dart}.

\subpara{Conditioning on target features.}
We concatenate the hidden states of several target layers and project them to the hidden dimension of the draft. At the first proposal position, this feature, taken from the confirmed prefix, is combined with the embedding of the last confirmed token. The subsequent positions use mask embeddings and a shared learned feature embedding. The draft processes these inputs jointly while retaining the target context keys and values of DSpark. The input thus carries context beyond token identity, which helps the draft account for similar motions that are represented by different tokens.

\subpara{Feature regression.}
We keep the CE term of \cref{eq:dspark_losses} and replace the TV term by a regression of target features, which moves the alignment from token distributions to hidden representations. Following EAGLE~\cite{eagle}, we match the target feature before the output head with a Smooth L1 loss. Let $h_i^t$ be the target feature that predicts $x_i$, $\widehat h_i$ its draft counterpart, and $\ell^{\mathrm{feat}}_i=\operatorname{SmoothL1}(\widehat h_i,h_i^t)$, averaged over the hidden dimensions:
\begin{equation}
\mathcal{L}
= \sum_i w_i\left(\lambda\,\ell^{\mathrm{CE}}_i + (1-\lambda)\,\ell^{\mathrm{feat}}_i\right),
\label{eq:feature_distillation}
\end{equation}
where $w_i$ and $\lambda$ are those of \cref{eq:dspark_losses}. The CE term encourages exact token agreement, and the regression term aligns the continuous representations from which the tokens are predicted.

\subsection{Continuation Reuse}
\label{sec:tree_fusion}

The purpose of continuation reuse is to exploit predictions that the target has already computed and that standard decoding discards. When a drafted token is rejected, the target has still predicted what would follow it. If the rejected token and the token the target wanted encode similar motions, what follows is likely to be the same, and it can be proposed in the next cycle (\cref{fig:tree_fusion}).

\subpara{Constructing draft trees.}
Unlike DSpark, which proposes a single chain, we construct a draft tree. Following prior work~\cite{chains2trees, zhou2026treespark}, we compute the base logits $U_i$ for all proposal positions once and apply the Markov correction of \cref{eq:markov} using each parent token, giving each branch its own candidate distribution without additional backbone passes. Candidates are expanded and nodes are retained as described in \cref{sec:preliminaries}.

\begin{figure}[t]
\centering
    \includegraphics[width=\linewidth]{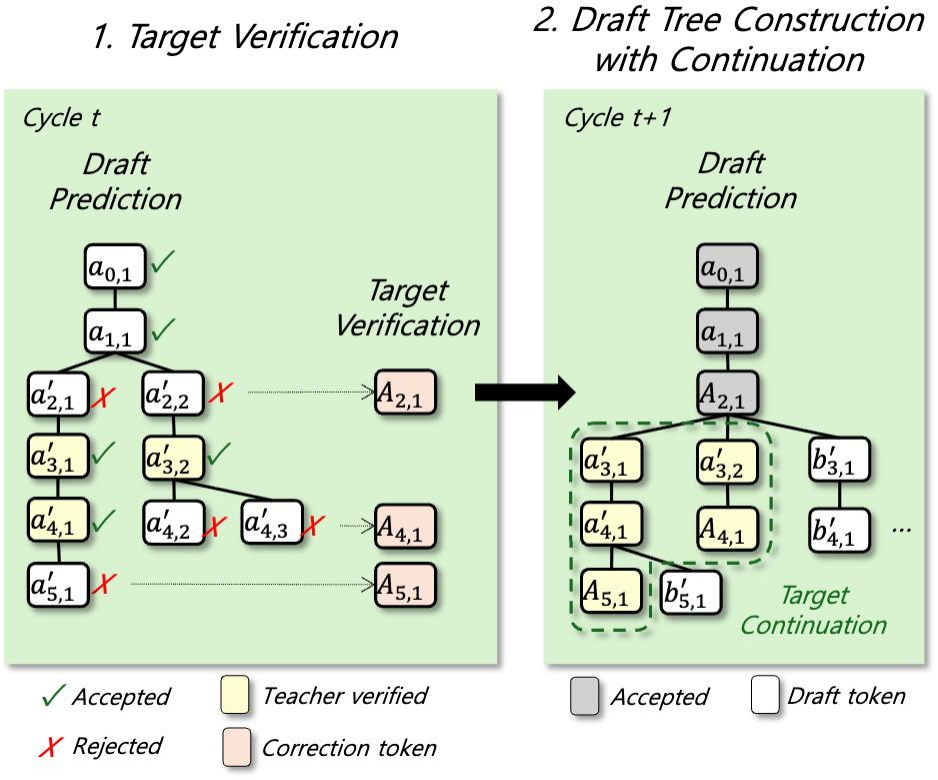}
    \figcapabove
    \caption{\textbf{Illustration of continuation reuse.} During verification in cycle $t$, the target also predicts along rejected branches. VSpark reuses them to build the tree for cycle $t+1$, where they are re-verified.}
    \label{fig:tree_fusion}
    \figcapbelow
\end{figure}

\subpara{Extracting continuations.}
Consider a cycle in which the tree proposes a token $a$ at the first position where no child matches, while the target predicts $b$. The correction token $b$ is appended to the confirmed prefix, and standard decoding discards the subtree under $a$. The same verification pass, however, has also produced the target's prediction for the token that follows $a$, say $c$. If $c$ is itself a child of $a$ in the tree, the target's prediction after $c$ is available as well, and so on. We follow these predictions through matching children and keep the final predicted token when no child matches; extraction also stops at an end-of-sequence token or at a length limit. The resulting sequence $(c,\dots)$ is a continuation: what the target would predict if $a$ had been accepted. We collect one continuation from each rejected child at that position. Continuations require no additional computation by the target.

\subpara{Promoting continuations.}
In the next cycle, the confirmed prefix ends with $b$ instead of $a$, and we use the continuations to guide the draft tree built from this corrected prefix. Wherever a branch of the new tree coincides with the beginning of a continuation, the next token of that continuation is suggested as a child of the branch. Let $s_i(v)=\log p_{\mathcal M_d}(v\mid o,\ell,x_{<i})$ be the draft score of token $v$ at a branch and $\mathcal H_i$ the set of tokens suggested there. Before candidates are selected, we raise the score of each suggested token to the highest draft score under the same parent:
\begin{equation}
\widetilde s_i(v)=
\begin{cases}
\max_u s_i(u), & v\in\mathcal H_i,\\
s_i(v), & \text{otherwise}.
\end{cases}
\label{eq:fusion_scores}
\end{equation}
A suggested token thus ties with the draft's best candidate even if its probability is low. We accumulate scores along each path and merge the continuations into the tree, sharing nodes that already exist, inserting those that do not, and pruning to the node budget.

\subpara{Early stopping.}
Log probabilities are non-positive, and \cref{eq:fusion_scores} preserves this, so extending a path cannot increase its cumulative score. Once the node budget is filled, we stop expanding when the best frontier score falls below the lowest selected score, since no descendant can then improve the selected set.

\subpara{Exactness.}
Continuation reuse changes only which candidates are proposed. Following common practice for autoregressive VLAs, we use greedy decoding: the target verifies every candidate under its actual prefix and accepts only tokens matching its highest-probability prediction. Promoted continuations are verified again under the corrected prefix, and only the accepted path enters the confirmed prefix, ensuring lossless decoding.
\begin{table*}[t]
\centering
\captionsetup{width=\textwidth}
\caption{\textbf{Main results on four LIBERO suites.} VSpark achieves the highest acceptance length $\tau$ and speedup on all four VLAs and all suites.}
\label{tab:main}
\vspace{-10pt}
\setlength{\tabcolsep}{9.7pt}
\scriptsize
\resizebox{\linewidth}{!}{%
\begin{tabular}{c@{\hspace{3pt}}l*{10}{c}}
\toprule
& & \multicolumn{2}{c}{Goal} & \multicolumn{2}{c}{Spatial} & \multicolumn{2}{c}{Object} & \multicolumn{2}{c}{Long} & \multicolumn{2}{c}{Avg.} \\
\cmidrule(lr){3-4}\cmidrule(lr){5-6}\cmidrule(lr){7-8}\cmidrule(lr){9-10}\cmidrule(lr){11-12}
& Method & $\tau\uparrow$ & Speedup$\uparrow$ & $\tau\uparrow$ & Speedup$\uparrow$ & $\tau\uparrow$ & Speedup$\uparrow$ & $\tau\uparrow$ & Speedup$\uparrow$ & $\tau\uparrow$ & Speedup$\uparrow$ \\
\midrule
\multirow{5}{*}{\rotatebox[origin=c]{90}{\scalebox{0.85}{$\pi_0$-FAST}}} & EAGLE3 & 3.09 & $1.22\times$ & 3.12 & $1.24\times$ & 3.31 & $1.31\times$ & 3.17 & $1.27\times$ & 3.17 & $1.26\times$ \\
& P-EAGLE & 2.94 & $1.45\times$ & 2.97 & $1.49\times$ & 3.12 & $1.57\times$ & 3.00 & $1.48\times$ & 3.01 & $1.50\times$ \\
& DFlash & 2.33 & $1.39\times$ & 2.42 & $1.46\times$ & 2.44 & $1.46\times$ & 2.30 & $1.36\times$ & 2.37 & $1.41\times$ \\
& DFlash2 & 2.99 & $1.50\times$ & 3.14 & $1.56\times$ & 3.20 & $1.59\times$ & 2.96 & $1.47\times$ & 3.07 & $1.53\times$ \\
& DSpark & 3.08 & $1.57\times$ & 3.12 & $1.59\times$ & 3.31 & $1.65\times$ & 3.14 & $1.60\times$ & 3.16 & $1.60\times$ \\
\rowcolor{aliceblue}
& \textbf{VSpark (Ours)} & \textbf{3.86} & $\mathbf{1.91}\times$ & \textbf{3.99} & $\mathbf{1.98}\times$ & \textbf{4.03} & $\mathbf{1.99}\times$ & \textbf{3.86} & $\mathbf{1.85}\times$ & \textbf{3.94} & $\mathbf{1.93}\times$ \\
\midrule
\multirow{5}{*}{\rotatebox[origin=c]{90}{\scalebox{0.85}{StarVLA-FAST}}} & EAGLE3 & 2.46 & $1.26\times$ & 2.31 & $1.24\times$ & 2.88 & $1.41\times$ & 2.81 & $1.40\times$ & 2.61 & $1.33\times$ \\
& P-EAGLE & 2.13 & $1.30\times$ & 2.03 & $1.30\times$ & 2.44 & $1.40\times$ & 2.38 & $1.42\times$ & 2.25 & $1.36\times$ \\
& DFlash & 1.51 & $1.10\times$ & 1.44 & $1.09\times$ & 1.63 & $1.14\times$ & 1.61 & $1.17\times$ & 1.55 & $1.12\times$ \\
& DFlash2 & 1.68 & $1.13\times$ & 1.63 & $1.12\times$ & 1.87 & $1.17\times$ & 1.83 & $1.22\times$ & 1.75 & $1.16\times$ \\
& DSpark & 2.13 & $1.31\times$ & 2.00 & $1.30\times$ & 2.46 & $1.46\times$ & 2.43 & $1.48\times$ & 2.26 & $1.39\times$ \\
\rowcolor{aliceblue}
& \textbf{VSpark (Ours)} & \textbf{2.60} & $\mathbf{1.56}\times$ & \textbf{2.43} & $\mathbf{1.56}\times$ & \textbf{2.94} & $\mathbf{1.64}\times$ & \textbf{2.91} & $\mathbf{1.68}\times$ & \textbf{2.72} & $\mathbf{1.61}\times$ \\
\midrule
\multirow{5}{*}{\rotatebox[origin=c]{90}{\scalebox{0.85}{ActionCodec}}} & EAGLE3 & 3.92 & $1.58\times$ & 3.76 & $1.57\times$ & 4.26 & $1.73\times$ & 4.01 & $1.65\times$ & 3.99 & $1.63\times$ \\
& P-EAGLE & 3.73 & $1.88\times$ & 3.66 & $1.90\times$ & 4.09 & $2.07\times$ & 3.83 & $1.96\times$ & 3.83 & $1.95\times$ \\
& DFlash & 3.30 & $1.84\times$ & 3.16 & $1.83\times$ & 3.64 & $2.02\times$ & 3.35 & $1.90\times$ & 3.36 & $1.90\times$ \\
& DFlash2 & 3.59 & $1.80\times$ & 3.44 & $1.75\times$ & 3.81 & $1.91\times$ & 3.56 & $1.77\times$ & 3.60 & $1.81\times$ \\
& DSpark & 3.60 & $1.85\times$ & 3.52 & $1.91\times$ & 3.94 & $2.04\times$ & 3.69 & $1.91\times$ & 3.69 & $1.93\times$ \\
\rowcolor{aliceblue}
& \textbf{VSpark (Ours)} & \textbf{4.13} & $\mathbf{2.04}\times$ & \textbf{4.03} & $\mathbf{2.06}\times$ & \textbf{4.44} & $\mathbf{2.21}\times$ & \textbf{4.23} & $\mathbf{2.08}\times$ & \textbf{4.21} & $\mathbf{2.10}\times$ \\
\midrule
\multirow{5}{*}{\rotatebox[origin=c]{90}{\scalebox{0.85}{G0.5}}} & EAGLE3 & 2.32 & $0.83\times$ & 2.33 & $0.80\times$ & 2.60 & $0.88\times$ & 2.33 & $0.80\times$ & 2.39 & $0.83\times$ \\
& P-EAGLE & 2.37 & $1.08\times$ & 2.35 & $1.06\times$ & 2.64 & $1.16\times$ & 2.36 & $1.06\times$ & 2.43 & $1.09\times$ \\
& DFlash & 1.75 & $0.95\times$ & 1.73 & $0.93\times$ & 1.90 & $1.03\times$ & 1.73 & $0.94\times$ & 1.78 & $0.96\times$ \\
& DFlash2 & 1.94 & $0.89\times$ & 1.91 & $0.87\times$ & 2.13 & $0.93\times$ & 1.92 & $0.88\times$ & 1.98 & $0.89\times$ \\
& DSpark & 2.26 & $1.03\times$ & 2.27 & $1.03\times$ & 2.54 & $1.10\times$ & 2.26 & $1.06\times$ & 2.33 & $1.05\times$ \\
\rowcolor{aliceblue}
& \textbf{VSpark (Ours)} & \textbf{2.95} & $\mathbf{1.27}\times$ & \textbf{2.95} & $\mathbf{1.27}\times$ & \textbf{3.15} & $\mathbf{1.30}\times$ & \textbf{2.92} & $\mathbf{1.27}\times$ & \textbf{2.99} & $\mathbf{1.28}\times$ \\
\bottomrule
\end{tabular}%
}
\vspace{-15pt}
\end{table*}

\begin{table}[t]
\centering
\captionsetup{width=\linewidth}
\caption{\textbf{SimplerEnv results} with StarVLA-FAST. Tasks: spoon on towel, carrot on plate, stack cubes, eggplant in basket. Spd.: speedup.}
\label{tab:simplerenv}
\tabcapbelow
\setlength{\tabcolsep}{0.8pt}
\scriptsize
\resizebox{\linewidth}{!}{%
\begin{tabular}{l<{\hspace{4pt}}*{10}{c}}
\toprule
& \multicolumn{2}{c}{Spoon}
& \multicolumn{2}{c}{Carrot}
& \multicolumn{2}{c}{Stack}
& \multicolumn{2}{c}{Eggplant}
& \multicolumn{2}{c}{Avg.} \\
\cmidrule(lr){2-3}\cmidrule(lr){4-5}
\cmidrule(lr){6-7}\cmidrule(lr){8-9}
\cmidrule(lr){10-11}
Method
& $\tau\uparrow$ & Spd.$\uparrow$
& $\tau\uparrow$ & Spd.$\uparrow$
& $\tau\uparrow$ & Spd.$\uparrow$
& $\tau\uparrow$ & Spd.$\uparrow$
& $\tau\uparrow$ & Spd.$\uparrow$ \\
\midrule
EAGLE3  & 3.89 & $1.69\times$ & 3.49 & $1.59\times$ & 3.68 & $1.64\times$ & 3.84 & $1.66\times$ & 3.72 & $1.65\times$ \\
P-EAGLE & 3.37 & $1.70\times$ & 2.95 & $1.53\times$ & 3.08 & $1.57\times$ & 3.26 & $1.62\times$ & 3.16 & $1.61\times$ \\
DFlash  & 2.85 & $1.51\times$ & 2.52 & $1.40\times$ & 2.59 & $1.41\times$ & 2.62 & $1.38\times$ & 2.64 & $1.42\times$ \\
DFlash2 & 3.21 & $1.62\times$ & 2.89 & $1.51\times$ & 2.97 & $1.51\times$ & 3.00 & $1.50\times$ & 3.02 & $1.54\times$ \\
DSpark  & 3.70 & $1.83\times$ & 3.31 & $1.72\times$ & 3.39 & $1.71\times$ & 3.50 & $1.74\times$ & 3.48 & $1.75\times$ \\
\rowcolor{aliceblue}
\textbf{VSpark}
& \textbf{4.01} & $\mathbf{1.99}\times$
& \textbf{3.87} & $\mathbf{1.96}\times$
& \textbf{3.81} & $\mathbf{1.93}\times$
& \textbf{3.94} & $\mathbf{1.92}\times$
& \textbf{3.90} & $\mathbf{1.95}\times$ \\
\bottomrule
\end{tabular}%
}
\vspace{-10pt}
\end{table}

\begin{table*}[t]
\centering
\captionsetup{width=\textwidth}
\caption{\textbf{Main results on LIBERO-Plus and LIBERO-Pro.}
Both benchmarks evaluate VLAs under diverse variations in visual appearance, object configurations, robot states, language instructions, and task conditions.
VSpark consistently achieves the highest acceptance length and inference speedup across all eight suites and four VLAs.
Avg. denotes the arithmetic mean across the eight suites.}
\label{tab:main_plus_pro}
\tabcapbelow
\setlength{\tabcolsep}{2.7pt}
\resizebox{\linewidth}{!}{%
\begin{tabular}{c@{\hspace{2pt}}l*{18}{c}}
\toprule
& & \multicolumn{8}{c}{LIBERO-Plus} & \multicolumn{8}{c}{LIBERO-Pro} & \multicolumn{2}{c}{} \\
\cmidrule(lr){3-10}\cmidrule(lr){11-18}
& & \multicolumn{2}{c}{Goal} & \multicolumn{2}{c}{Spatial} & \multicolumn{2}{c}{Object} & \multicolumn{2}{c}{Long}
& \multicolumn{2}{c}{Goal} & \multicolumn{2}{c}{Spatial} & \multicolumn{2}{c}{Object} & \multicolumn{2}{c}{Long} & \multicolumn{2}{c}{Avg.} \\
\cmidrule(lr){3-4}\cmidrule(lr){5-6}\cmidrule(lr){7-8}\cmidrule(lr){9-10}\cmidrule(lr){11-12}\cmidrule(lr){13-14}\cmidrule(lr){15-16}\cmidrule(lr){17-18}\cmidrule(lr){19-20}
& Method & $\tau\uparrow$ & Speedup$\uparrow$ & $\tau\uparrow$ & Speedup$\uparrow$ & $\tau\uparrow$ & Speedup$\uparrow$ & $\tau\uparrow$ & Speedup$\uparrow$ & $\tau\uparrow$ & Speedup$\uparrow$ & $\tau\uparrow$ & Speedup$\uparrow$ & $\tau\uparrow$ & Speedup$\uparrow$ & $\tau\uparrow$ & Speedup$\uparrow$ & $\tau\uparrow$ & Speedup$\uparrow$ \\
\midrule
\multirow{5}{*}{\rotatebox[origin=c]{90}{\scalebox{0.85}{$\pi_0$-FAST}}} & EAGLE3 & 2.74 & $1.00\times$ & 2.78 & $1.01\times$ & 2.94 & $1.07\times$ & 2.78 & $1.00\times$ & 2.95 & $1.45\times$ & 3.03 & $1.48\times$ & 3.29 & $1.58\times$ & 3.06 & $1.46\times$ & 2.95 & $1.26\times$ \\
& P-EAGLE & 2.60 & $1.40\times$ & 2.63 & $1.39\times$ & 2.77 & $1.49\times$ & 2.64 & $1.33\times$ & 2.79 & $1.69\times$ & 2.88 & $1.76\times$ & 3.09 & $1.81\times$ & 2.89 & $1.75\times$ & 2.79 & $1.58\times$ \\
& DFlash & 1.92 & $1.23\times$ & 1.94 & $1.25\times$ & 1.92 & $1.25\times$ & 1.89 & $1.18\times$ & 2.25 & $1.68\times$ & 2.36 & $1.72\times$ & 2.39 & $1.73\times$ & 2.26 & $1.66\times$ & 2.12 & $1.46\times$ \\
& DFlash2 & 2.46 & $1.32\times$ & 2.53 & $1.37\times$ & 2.62 & $1.42\times$ & 2.44 & $1.29\times$ & 2.73 & $1.68\times$ & 2.96 & $1.78\times$ & 3.07 & $1.80\times$ & 2.80 & $1.67\times$ & 2.70 & $1.54\times$ \\
& DSpark & 2.70 & $1.46\times$ & 2.70 & $1.46\times$ & 2.89 & $1.54\times$ & 2.75 & $1.44\times$ & 2.94 & $1.82\times$ & 3.02 & $1.85\times$ & 3.28 & $1.93\times$ & 3.06 & $1.83\times$ & 2.92 & $1.67\times$ \\
\rowcolor{aliceblue}
& \textbf{VSpark} & \textbf{3.32} & $\mathbf{1.78}\times$ & \textbf{3.34} & $\mathbf{1.75}\times$ & \textbf{3.49} & $\mathbf{1.78}\times$ & \textbf{3.33} & $\mathbf{1.65}\times$ & \textbf{3.62} & $\mathbf{2.14}\times$ & \textbf{3.78} & $\mathbf{2.24}\times$ & \textbf{3.92} & $\mathbf{2.21}\times$ & \textbf{3.71} & $\mathbf{2.13}\times$ & \textbf{3.56} & $\mathbf{1.96}\times$ \\
\midrule
\multirow{5}{*}{\rotatebox[origin=c]{90}{\scalebox{0.85}{StarVLA-FAST}}} & EAGLE3 & 2.10 & $1.12\times$ & 2.03 & $1.07\times$ & 2.39 & $1.24\times$ & 2.27 & $1.17\times$ & 2.25 & $1.44\times$ & 2.10 & $1.36\times$ & 2.91 & $1.71\times$ & 2.54 & $1.56\times$ & 2.32 & $1.33\times$ \\
& P-EAGLE & 1.87 & $1.18\times$ & 1.82 & $1.13\times$ & 2.08 & $1.27\times$ & 1.99 & $1.24\times$ & 1.99 & $1.51\times$ & 1.88 & $1.42\times$ & 2.46 & $1.74\times$ & 2.19 & $1.60\times$ & 2.04 & $1.39\times$ \\
& DFlash & 1.34 & $1.01\times$ & 1.32 & $0.98\times$ & 1.44 & $1.06\times$ & 1.41 & $1.03\times$ & 1.41 & $1.25\times$ & 1.35 & $1.21\times$ & 1.57 & $1.36\times$ & 1.49 & $1.31\times$ & 1.42 & $1.15\times$ \\
& DFlash2 & 1.51 & $1.00\times$ & 1.49 & $0.96\times$ & 1.67 & $1.07\times$ & 1.61 & $1.04\times$ & 1.57 & $1.25\times$ & 1.52 & $1.21\times$ & 1.80 & $1.38\times$ & 1.69 & $1.33\times$ & 1.61 & $1.16\times$ \\
& DSpark & 1.79 & $1.14\times$ & 1.72 & $1.09\times$ & 2.00 & $1.23\times$ & 1.95 & $1.21\times$ & 1.98 & $1.50\times$ & 1.85 & $1.43\times$ & 2.50 & $1.77\times$ & 2.22 & $1.61\times$ & 2.00 & $1.37\times$ \\
\rowcolor{aliceblue}
& \textbf{VSpark} & \textbf{2.22} & $\mathbf{1.36}\times$ & \textbf{2.13} & $\mathbf{1.31}\times$ & \textbf{2.48} & $\mathbf{1.48}\times$ & \textbf{2.39} & $\mathbf{1.42}\times$ & \textbf{2.39} & $\mathbf{1.75}\times$ & \textbf{2.22} & $\mathbf{1.65}\times$ & \textbf{2.95} & $\mathbf{1.98}\times$ & \textbf{2.65} & $\mathbf{1.87}\times$ & \textbf{2.43} & $\mathbf{1.60}\times$ \\
\midrule
\multirow{5}{*}{\rotatebox[origin=c]{90}{\scalebox{0.85}{ActionCodec}}} & EAGLE3 & 2.95 & $1.35\times$ & 2.80 & $1.31\times$ & 3.24 & $1.51\times$ & 3.17 & $1.46\times$ & 3.36 & $1.49\times$ & 3.50 & $1.60\times$ & 4.24 & $1.81\times$ & 3.66 & $1.57\times$ & 3.36 & $1.51\times$ \\
& P-EAGLE & 2.85 & $1.64\times$ & 2.67 & $1.61\times$ & 3.09 & $1.78\times$ & 3.04 & $1.72\times$ & 3.20 & $1.70\times$ & 3.35 & $1.89\times$ & 4.12 & $2.11\times$ & 3.50 & $1.81\times$ & 3.23 & $1.78\times$ \\
& DFlash & 2.39 & $1.53\times$ & 2.25 & $1.47\times$ & 2.57 & $1.68\times$ & 2.56 & $1.64\times$ & 2.75 & $1.65\times$ & 2.91 & $1.79\times$ & 3.55 & $2.03\times$ & 3.03 & $1.78\times$ & 2.75 & $1.70\times$ \\
& DFlash2 & 2.55 & $1.48\times$ & 2.39 & $1.38\times$ & 2.78 & $1.56\times$ & 2.75 & $1.55\times$ & 2.93 & $1.55\times$ & 3.10 & $1.69\times$ & 3.81 & $1.94\times$ & 3.25 & $1.68\times$ & 2.95 & $1.60\times$ \\
& DSpark & 2.72 & $1.63\times$ & 2.50 & $1.48\times$ & 2.91 & $1.70\times$ & 2.86 & $1.70\times$ & 3.07 & $1.69\times$ & 3.24 & $1.83\times$ & 3.93 & $2.08\times$ & 3.39 & $1.81\times$ & 3.08 & $1.74\times$ \\
\rowcolor{aliceblue}
& \textbf{VSpark} & \textbf{3.37} & $\mathbf{1.89}\times$ & \textbf{3.15} & $\mathbf{1.81}\times$ & \textbf{3.57} & $\mathbf{1.99}\times$ & \textbf{3.55} & $\mathbf{1.98}\times$ & \textbf{3.76} & $\mathbf{2.01}\times$ & \textbf{3.84} & $\mathbf{2.14}\times$ & \textbf{4.58} & $\mathbf{2.35}\times$ & \textbf{4.06} & $\mathbf{2.12}\times$ & \textbf{3.73} & $\mathbf{2.04}\times$ \\
\midrule
\multirow{5}{*}{\rotatebox[origin=c]{90}{\scalebox{0.85}{G0.5}}} & EAGLE3 & 1.98 & $0.71\times$ & 2.02 & $0.74\times$ & 2.22 & $0.80\times$ & 2.08 & $0.76\times$ & 2.21 & $0.87\times$ & 2.25 & $0.88\times$ & 2.57 & $0.97\times$ & 2.29 & $0.89\times$ & 2.20 & $0.83\times$ \\
& P-EAGLE & 1.92 & $0.90\times$ & 1.95 & $0.97\times$ & 2.09 & $1.00\times$ & 2.01 & $0.98\times$ & 2.19 & $1.10\times$ & 2.25 & $1.12\times$ & 2.56 & $1.24\times$ & 2.27 & $1.14\times$ & 2.15 & $1.06\times$ \\
& DFlash & 1.55 & $0.85\times$ & 1.56 & $0.89\times$ & 1.66 & $0.95\times$ & 1.61 & $0.92\times$ & 1.64 & $0.99\times$ & 1.66 & $1.00\times$ & 1.83 & $1.08\times$ & 1.68 & $1.02\times$ & 1.65 & $0.96\times$ \\
& DFlash2 & 1.65 & $0.79\times$ & 1.65 & $0.80\times$ & 1.79 & $0.85\times$ & 1.72 & $0.83\times$ & 1.80 & $0.92\times$ & 1.83 & $0.92\times$ & 2.06 & $1.01\times$ & 1.86 & $0.94\times$ & 1.79 & $0.88\times$ \\
& DSpark & 1.95 & $0.90\times$ & 2.01 & $0.95\times$ & 2.21 & $1.03\times$ & 2.07 & $0.97\times$ & 2.11 & $1.08\times$ & 2.16 & $1.10\times$ & 2.46 & $1.21\times$ & 2.18 & $1.11\times$ & 2.14 & $1.04\times$ \\
\rowcolor{aliceblue}
& \textbf{VSpark} & \textbf{2.61} & $\mathbf{1.19}\times$ & \textbf{2.66} & $\mathbf{1.21}\times$ & \textbf{2.80} & $\mathbf{1.24}\times$ & \textbf{2.70} & $\mathbf{1.25}\times$ & \textbf{2.72} & $\mathbf{1.32}\times$ & \textbf{2.80} & $\mathbf{1.35}\times$ & \textbf{2.98} & $\mathbf{1.39}\times$ & \textbf{2.81} & $\mathbf{1.35}\times$ & \textbf{2.76} & $\mathbf{1.29}\times$ \\
\bottomrule
\end{tabular}%
}
\vspace{-10pt}
\end{table*}

\section{Experiments}
\label{sec:experiments}

\subsection{Experimental Setup}

\subpara{Target VLAs.}
We accelerate four publicly released autoregressive VLAs that cover two families of action tokenizer. $\pi_0$-FAST and StarVLA-FAST~\cite{community2026starvla} use FAST tokenization, and ActionCodec-AR and G0.5 use learned vector-quantized tokenizers. The main results cover all four, while the ablations and analyses use one model per family, $\pi_0$-FAST and ActionCodec.

\subpara{Speculative decoding baselines.}
We compare VSpark with five speculative decoding methods: EAGLE-3~\cite{eagle3}, P-EAGLE~\cite{peagle}, DFlash~\cite{dflash}, DFlash2~\cite{dflash2}, and DSpark~\cite{dspark}. Every method accelerates the same frozen targets, so the methods differ only in their draft. We adapt each to action tokens and extend the parallel drafters to trees, so that all are verified alike.

\subpara{Benchmarks.}
LIBERO comprises four suites, Goal, Spatial, Object, and Long, each with ten tasks and 50 initial states per task, giving 500 episodes per model and suite. LIBERO-Plus and LIBERO-Pro evaluate the same suites under distribution shift: LIBERO-Plus perturbs camera viewpoints, robot initial states, language instructions, lighting, backgrounds, sensor noise, and object layouts, and LIBERO-Pro varies objects, positions, language semantics, tasks, and environments. SimplerEnv (WidowX) evaluates a different embodiment, using StarVLA-FAST trained on BridgeData V2 on four tasks with 24 episodes each. We follow the official protocol of each benchmark.

\subpara{Metrics.}
We report the acceptance length $\tau$, the average number of tokens committed per verification cycle including the correction or additional token, and the speedup, the autoregressive decoding time of the target divided by the speculative decoding time in the same setting. Avg. denotes the arithmetic mean over the suites or tasks of a benchmark.

\subpara{Implementation details.}
Unless otherwise specified, all drafts have one layer and are trained for ten epochs with AdamW~\cite{adamw}, a batch size of 16, a peak learning rate of $1\times10^{-4}$, $\beta=(0.9,0.95)$, no weight decay, warmup over the first 5\% of training followed by linear decay, and gradient clipping at 0.5. All methods receive the same number of optimizer updates, and we report the final-epoch checkpoint. Following DSpark, we set $\lambda=0.1$. 
We retain EAGLE-3's tree drafting and extend all other baselines to trees. DSpark and VSpark reuse base logits with Markov corrections; the others require additional draft-model passes for tree expansion.
Decoding is greedy with exact verification, on draft trees with a maximum depth of seven, a budget of 60 nodes, and eight candidates per parent. Generation-error episodes and token-limit chunks are excluded from all results. All experiments run on a single NVIDIA RTX 4090 GPU.

\subsection{Experimental Results}
\label{sec:experimental_results}

\subpara{Results on LIBERO.}
\cref{tab:main} compares VSpark with the baselines on the four LIBERO suites. VSpark achieves the highest acceptance length and speedup on every model and suite, with average speedups of $1.93\times$, $1.61\times$, $2.10\times$, and $1.28\times$ on $\pi_0$-FAST, StarVLA-FAST, ActionCodec, and G0.5. Relative to DSpark, on which it is built, the average acceptance length rises from 3.16 to 3.94, 2.26 to 2.72, 3.69 to 4.21, and 2.33 to 2.99. Two observations follow. First, a longer acceptance length does not by itself make decoding faster: on $\pi_0$-FAST, EAGLE-3 accepts as many tokens as DSpark (3.17 against 3.16) but reaches only $1.26\times$ against $1.60\times$, since its draft generates tokens autoregressively rather than in parallel. Second, G0.5's linear attention limits parallel tree verification: three baselines are slower than autoregressive decoding and the best reaches $1.09\times$, whereas VSpark reaches $1.28\times$. The improvement thus holds for both frequency-based and learned vector-quantized tokenizers.

\begin{table}[t]
\centering
\caption{\textbf{Embedding-based relaxed verification on LIBERO} with $\pi_0$-FAST and DSpark drafts, averaged over four suites. A draft token is accepted when its cosine distance from the target token embedding is at most the given distance. Relaxed ratio is the fraction of accepted draft tokens that differ from the target prediction.}
\label{tab:embedding_relaxed}
\vspace{-10pt}
\setlength{\tabcolsep}{9.4pt}
\scriptsize
\resizebox{\linewidth}{!}{%
\begin{tabular}{ccccc}
\toprule
Distance & Relaxed ratio
& $\tau\uparrow$ & Speedup$\uparrow$ & SR (\%)$\uparrow$ \\
\midrule
-   & 0.00\%  & 3.16 & $1.60\times$ & 91.7 \\
\midrule
0.011 & 2.05\%  & 3.18 & $1.62\times$ & 91.7 \\
0.012 & 12.70\% & 3.31 & $1.67\times$ & 91.0 \\
0.013 & 23.72\% & 3.47 & $1.74\times$ & 89.2 \\
0.015 & 35.79\% & 3.70 & $1.82\times$ & 83.8 \\
0.020 & 45.97\% & 4.02 & $1.90\times$ & 71.4 \\
0.025 & 47.56\% & 4.08 & $1.93\times$ & 70.5 \\
\bottomrule
\end{tabular}%
}
\vspace{-10pt}
\end{table}
\begin{figure}[t]
\centering
\includegraphics[width=\linewidth]{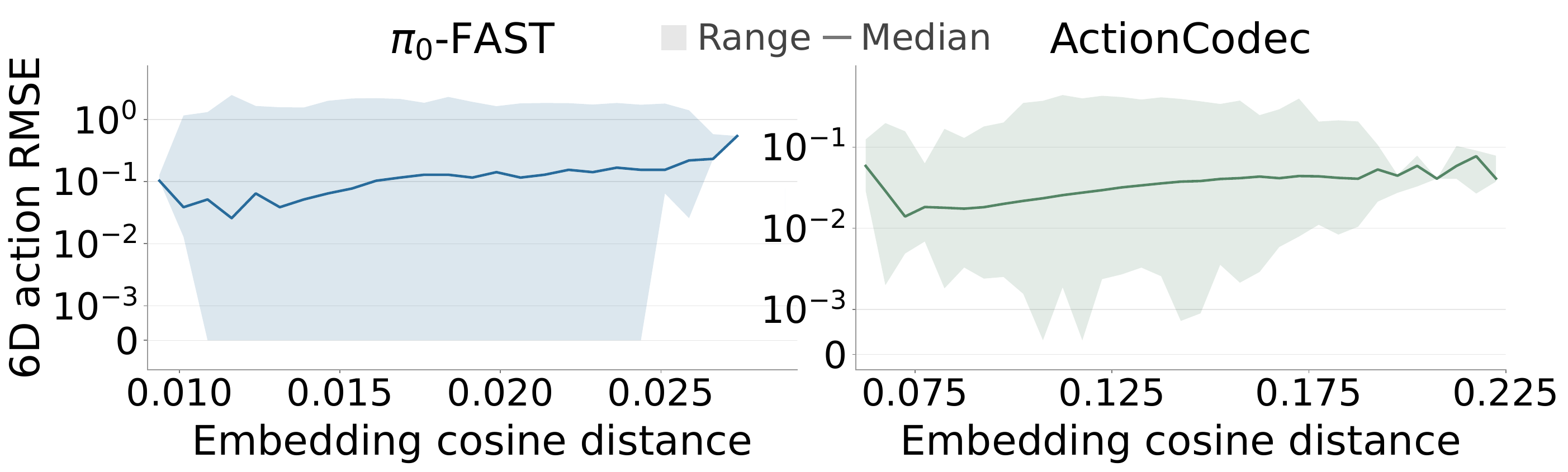}
    \figcapabove
    \caption{\textbf{Token embedding distance versus action discrepancy} for $\pi_0$-FAST (left) and ActionCodec (right). Lines show the median and shaded regions the range of the 6D action RMSE at each distance. Similar distances span a wide range of action errors.}
    \label{fig:action_distance}
    \vspace{-20pt}
\end{figure}

\subpara{Results on SimplerEnv.}
\cref{tab:simplerenv} evaluates a different embodiment (WidowX) and a model trained on real-world robot data. VSpark achieves an average acceptance length of 3.90 and a speedup of $1.95\times$, against 3.48 and $1.75\times$ for DSpark, and it is the fastest method on each of the four tasks, with speedups between $1.92\times$ and $1.99\times$. The benefit of VSpark therefore carries over to another embodiment and environment.

\subpara{Results on LIBERO-Plus and LIBERO-Pro.}
\cref{tab:main_plus_pro} repeats the comparison under the distribution shifts of LIBERO-Plus and LIBERO-Pro. VSpark again achieves the highest $\tau$ and speedup on all eight suites for all four models, with average speedups of $1.96\times$, $1.60\times$, $2.04\times$, and $1.29\times$. The acceptance length drops for every method under shift: that of DSpark on $\pi_0$-FAST falls from 3.16 on LIBERO to 2.92. The margin of VSpark over DSpark is nevertheless preserved (3.56 against 2.92 on $\pi_0$-FAST, and 3.73 against 3.08 on ActionCodec), so its benefit is not tied to the standard setting.

\subsection{Analysis}
\label{sec:analysis}

\subpara{Token embedding similarity versus action similarity.}
Relaxed verification needs a measure of how close two tokens are; for compressed tokenizers, where action distance requires complete token sequences, the natural candidate is the distance between token embeddings, which \cref{fig:action_distance} tests. 
For rejected candidates proposed by DSpark under the same prefix, we substitute one token into the target's sequence, keep the remaining tokens fixed, and decode the action. We measure embedding distance as one minus the cosine similarity between candidate and target token embeddings. Ranking the candidates by embedding distance disagrees with ranking them by action distance for 30.62\% and 31.47\% of valid pairs on $\pi_0$-FAST and ActionCodec, and 18.15\% of the substitutions on $\pi_0$-FAST do not even yield a valid coefficient sequence. \cref{tab:embedding_relaxed} shows that a 0.004 threshold increase (0.011 to 0.015) raises the relaxed ratio from 2.05\% to 35.79\% and speedup from $1.62\times$ to $1.82\times$, but lowers success from 91.7\% to 83.8\%. This suggests densely clustered draft candidates, not reliable action similarity. At 0.020, speedup rises from $1.60\times$ to $1.90\times$ but success falls to 71.4\%. Matching the $1.93\times$ of VSpark through relaxation reduces success to 70.5\%, whereas VSpark retains exact verification.

\begin{table}[t]
\centering
\caption{\textbf{Next-token agreement after rejection} (\%), over four LIBERO suites. Draft and Target: how often the next token each predicts after the rejected token matches the target's next token after the correction. $\Delta$: their difference in percentage points.}
\vspace{0.2em}
\label{tab:continuation_survival}
\tabcapbelow
\setlength{\tabcolsep}{6.1pt}
\scriptsize
\resizebox{\linewidth}{!}{%
\begin{tabular}{l*{6}{c}}
\toprule
& \multicolumn{3}{c}{$\pi_0$-FAST}
& \multicolumn{3}{c}{ActionCodec} \\
\cmidrule(lr){2-4}\cmidrule(lr){5-7}
Reject Pos.
& Draft & Target & $\Delta$
& Draft & Target & $\Delta$ \\
\midrule
1   & 24.70 & 39.50 & +14.80
    & 35.55 & 54.24 & +18.69 \\
2   & 35.20 & 45.96 & +10.76
    & 38.54 & 60.05 & +21.51 \\
3   & 28.20 & 55.12 & +26.92
    & 45.14 & 65.48 & +20.35 \\
4   & 29.26 & 53.40 & +24.14
    & 32.68 & 47.75 & +15.07 \\
5--7 & 46.57 & 56.42 & +9.85
     & 36.22 & 54.35 & +18.13 \\
\midrule
\textbf{Overall}
& 27.05 & \textbf{45.38} & \textbf{+18.33}
& 36.35 & \textbf{54.88} & \textbf{+18.52} \\
\bottomrule
\end{tabular}%
}
\vspace{-15pt}
\end{table}
\begin{table*}[t]
\centering
\caption{\textbf{Effect of verification strategy and draft depth.}
Tree verification consistently improves acceptance length and speedup over chain verification.
Increasing draft depth to five layers generally improves acceptance length but reduces speedup.}
\label{tab:draft_depth_ablation}
\vspace{-10pt}
\setlength{\tabcolsep}{6.8pt}
\scriptsize
\resizebox{\textwidth}{!}{%
\begin{tabular}{l*{12}{c}}
\toprule
& \multicolumn{6}{c}{$\pi_0$-FAST}
& \multicolumn{6}{c}{ActionCodec} \\
\cmidrule(lr){2-7}\cmidrule(lr){8-13}
& \multicolumn{4}{c}{1 Layer}
& \multicolumn{2}{c}{5 Layers}
& \multicolumn{4}{c}{1 Layer}
& \multicolumn{2}{c}{5 Layers} \\
\cmidrule(lr){2-5}\cmidrule(lr){6-7}
\cmidrule(lr){8-11}\cmidrule(lr){12-13}
& \multicolumn{2}{c}{Chain}
& \multicolumn{2}{c}{Tree}
& \multicolumn{2}{c}{Tree}
& \multicolumn{2}{c}{Chain}
& \multicolumn{2}{c}{Tree}
& \multicolumn{2}{c}{Tree} \\
\cmidrule(lr){2-3}\cmidrule(lr){4-5}
\cmidrule(lr){6-7}\cmidrule(lr){8-9}
\cmidrule(lr){10-11}\cmidrule(lr){12-13}
Method
& $\tau\uparrow$ & Speedup$\uparrow$
& $\tau\uparrow$ & Speedup$\uparrow$
& $\tau\uparrow$ & Speedup$\uparrow$
& $\tau\uparrow$ & Speedup$\uparrow$
& $\tau\uparrow$ & Speedup$\uparrow$
& $\tau\uparrow$ & Speedup$\uparrow$ \\
\midrule
EAGLE3
& 2.28 & $1.04\times$
& 3.17 & $\mathbf{1.26}\times$
& \textbf{3.35} & $0.70\times$
& 2.42 & $1.13\times$
& 3.99 & $\mathbf{1.63}\times$
& \textbf{4.11} & $0.98\times$ \\
P-EAGLE
& 1.99 & $1.16\times$
& 3.01 & $\mathbf{1.50}\times$
& \textbf{3.11} & $1.17\times$
& 2.36 & $1.36\times$
& 3.83 & $\mathbf{1.95}\times$
& \textbf{4.04} & $1.67\times$ \\
DFlash
& 1.55 & $1.05\times$
& 2.37 & $\mathbf{1.41}\times$
& \textbf{2.73} & $1.20\times$
& 2.17 & $1.37\times$
& \textbf{3.36} & $\mathbf{1.90}\times$
& 3.27 & $1.51\times$ \\
DFlash2
& 1.87 & $1.10\times$
& \textbf{3.07} & $\mathbf{1.53}\times$
& 2.96 & $1.05\times$
& 2.26 & $1.26\times$
& 3.60 & $\mathbf{1.81}\times$
& \textbf{4.01} & $1.48\times$ \\
DSpark
& 2.12 & $1.26\times$
& 3.16 & $\mathbf{1.60}\times$
& \textbf{3.35} & $1.33\times$
& 2.31 & $1.37\times$
& 3.69 & $\mathbf{1.93}\times$
& \textbf{3.93} & $1.66\times$ \\
\midrule
\rowcolor{aliceblue}
\textbf{VSpark}
& 2.28 & $1.31\times$
& \textbf{3.94} & $\mathbf{1.93}\times$
& 3.92 & $1.45\times$
& 2.46 & $1.44\times$
& 4.21 & $\mathbf{2.10}\times$
& \textbf{4.33} & $1.69\times$ \\
\bottomrule
\end{tabular}%
}
\vspace{-15pt}
\end{table*}

\subpara{Utility of rejected branch predictions.}
Continuation reuse assumes that what the target predicts after a rejected token is still useful after the correction. To test this, \cref{tab:continuation_survival} compares, at each rejection, the next token that the target predicts after the rejected token with the one it predicts after the correction. They agree in 45.38\% and 54.88\% of cases on $\pi_0$-FAST and ActionCodec, which is 18.33 and 18.52 percentage points more often than the next-token prediction of the draft, and the target is the more accurate of the two at every rejection position. A rejected token thus leaves the continuation unchanged about half of the time, consistent with rejected and corrected tokens encoding similar motions. \cref{fig:fusion_depth} shows the effect inside the tree. Among candidates whose entire parent path has been accepted, those promoted by continuation reuse are accepted far more often than the others at every depth, for example 70.18\% and 71.40\% against 14.32\% and 11.37\% at depth two, and 91.04\% and 85.96\% against 50.54\% and 47.53\% at depth six. Predictions from rejected branches are therefore reliable candidates, which is why reusing them lengthens the accepted sequence.

\begin{figure}[t]
\centering
\includegraphics[width=\linewidth]{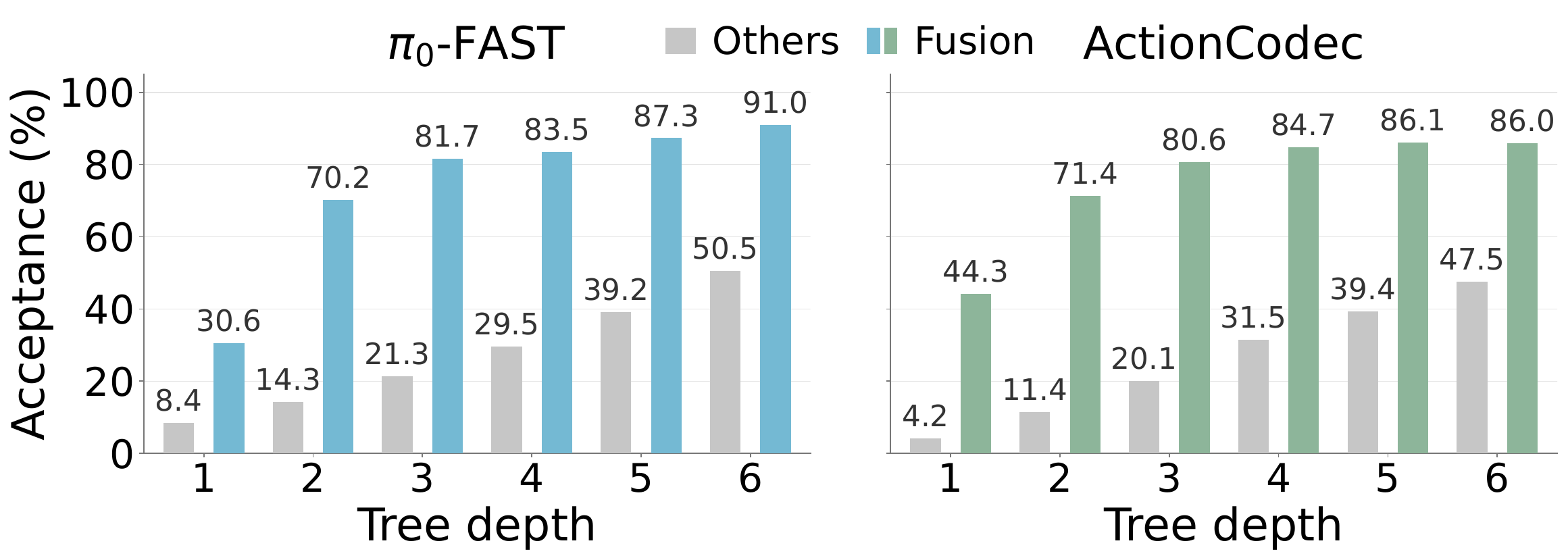}
    \figcapabove
    \caption{\textbf{Conditional token acceptance in VSpark} for $\pi_0$-FAST (left) and ActionCodec (right): acceptance rate of candidates promoted by continuation reuse (Fusion) and of the others, among candidates whose entire parent path has been accepted.}
\label{fig:fusion_depth}
    \vspace{-15pt}
\end{figure}

\subsection{Ablation Study}
\label{sec:ablation}

\subpara{Effect of verification strategy and draft depth.}
\cref{tab:draft_depth_ablation} compares verifying a single chain with verifying a tree, and one-layer with five-layer drafts. Trees help every method, as exact matching of action tokens is hard and a tree offers the target several candidates at each position. VSpark benefits the most, since continuation reuse operates only on trees: its $\tau$ rises from 2.28 to 3.94 on $\pi_0$-FAST and from 2.46 to 4.21 on ActionCodec, and its speedup from $1.31\times$ to $1.93\times$ and from $1.44\times$ to $2.10\times$. Deeper drafts, common in DFlash-based methods, do not pay off. With five layers, the acceptance length of VSpark changes little (3.94 to 3.92 and 4.21 to 4.33), while the cost of the draft lowers its speedup to $1.45\times$ and $1.69\times$, and EAGLE-3 becomes slower than autoregressive decoding. We therefore use one-layer drafts with tree verification.

\subpara{Component ablation.}
\cref{tab:component_ablation} measures the contribution of each design choice by adding them to DSpark one at a time. We start with continuation reuse, which acts only at inference and requires no retraining. It raises $\tau$ from 3.16 to 3.22 on $\pi_0$-FAST and from 3.69 to 4.04 on ActionCodec. Feature alignment raises it further to 3.87 and 4.10, 
adding action noise to rollouts raises it to 3.94 and 4.21, with speedups of $1.93\times$ and $2.10\times$.
Every component improves both models, but their shares differ: feature alignment accounts for most of the gain on $\pi_0$-FAST and continuation reuse for most of it on ActionCodec, while noise rollouts add a smaller, consistent improvement. The three choices are therefore complementary rather than interchangeable.

\begin{table}[t]
\centering
\caption{\textbf{Component ablation of VSpark.}
Each component progressively improves acceptance length and inference speedup on both VLAs.
Feature alignment provides the largest improvement on $\pi_0$-FAST, while continuation reuse contributes most on ActionCodec.
Combining all three components achieves speedups of $1.93\times$ and $2.10\times$, respectively.}
\vspace{0.1em}
\label{tab:component_ablation}
\tabcapbelow
\setlength{\tabcolsep}{8.7pt}
\scriptsize
\resizebox{\linewidth}{!}{%
\begin{tabular}{lcccc}
\toprule
& \multicolumn{2}{c}{$\pi_0$-FAST}
& \multicolumn{2}{c}{ActionCodec} \\
\cmidrule(lr){2-3}\cmidrule(lr){4-5}
Method
& $\tau \uparrow$ & Speedup $\uparrow$
& $\tau \uparrow$ & Speedup $\uparrow$ \\
\midrule
DSpark         & 3.16 & $1.60\times$ & 3.69 & $1.93\times$ \\
+ Continuation Reuse  & 3.22 & $1.69\times$ & 4.04 & $2.01\times$ \\
+ Feature Alignment     & 3.87 & $1.90\times$ & 4.10 & $2.07\times$ \\
\rowcolor{aliceblue}
+ Action Noise \textbf{(VSpark)} & \textbf{3.94} & $\mathbf{1.93}\times$
               & \textbf{4.21} & $\mathbf{2.10}\times$ \\
\bottomrule
\end{tabular}%
}
\vspace{-10pt}
\end{table}

\section{Conclusion}
We presented VSpark, a speculative decoding framework that provides lossless acceleration for autoregressive VLAs. Its premise is that the similarity between robot actions should be exploited when drafting rather than when verifying: calibrated-noise rollouts and feature alignment teach the draft more than token labels can, and continuation reuse recycles predictions that the target has already made along rejected branches, at no additional cost. With exact verification, VSpark is the fastest of six speculative decoders on every model and benchmark we tested, including targets on which most baselines are slower than autoregressive decoding. Despite these gains, VSpark has limitations that follow from its design. 
Hybrid attention has a cost: recurrent states in the linear attention layers of G0.5 complicate parallel tree verification, potentially limiting speedup despite improved acceptance.
Continuation reuse helps only when a rejection leaves the continuation unchanged, which holds for about half of the rejections and varies with the tokenizer.
Finally, VSpark helps a target trained on real-robot data (WidowX), but collecting noisy rollouts on a real robot without a simulator requires further study.
Improving parallel verification for hybrid attention architectures and collecting noise rollouts without a simulator remain directions for future work.

\section*{LLM Usage Statement}
We used Claude to improve the clarity and grammar of the manuscript and to assist with code refinement. The authors reviewed and revised the suggested edits, verified the resulting code, and take full responsibility for the final manuscript and implementation.

{
    \small
    \bibliographystyle{ieeenat_fullname}
    \bibliography{shortstrings,main}
}

  \savesuppcitationunit
\clearpage
\ifsubmission
\setcounter{page}{1}
\fi
\appendix

\renewcommand{\theHsection}{appendix.\Alph{section}}
\renewcommand{\theHsubsection}{appendix.\Alph{section}.\arabic{subsection}}

\counterwithin{figure}{section}
\counterwithin{table}{section}

\renewcommand{\thefigure}{\thesection.\arabic{figure}}
\renewcommand{\thetable}{\thesection.\arabic{table}}

\renewcommand{\cvprsect}[1]{\cvprsection[#1]{\texorpdfstring{\hskip -1em.~}{}#1}}
\renewcommand{\cvprsubsect}[1]{\cvprsubsection[#1]{\texorpdfstring{\hskip -1em.~}{}#1}}

\twocolumn[{
\begin{center}
    \Large \textbf{VSpark: Speculative Decoding for Autoregressive Vision-Language-Action Models} \\
    \vspace{0.5em}
    \large \textbf{Supplementary Material} \\
    \vspace{1.5em}
\end{center}
\begin{minipage}{\textwidth}
\titlecontents{lsection}[2em]{\addvspace{0.3em}\bfseries}
  {\contentslabel[\thecontentslabel.]{2em}}{}{\hfill\contentspage}
\titlecontents{lsubsection}[4.7em]{}
  {\contentslabel[\thecontentslabel.]{3.2em}}{}{\titlerule*[0.5pc]{.}\contentspage}
\startcontents[supp]
\hypersetup{linkcolor=black}
\printcontents[supp]{l}{1}{\setcounter{tocdepth}{2}}
\end{minipage}
\vspace{1em}
}]

\restoresuppcitationunit
\section{Experimental Details}
\label{sec:supp_details}

\subsection{Evaluation Protocols}
\label{sec:supp_protocols}
\cref{tab:supp_episodes} reports the number of episodes for each target model, decoding method, and evaluation setting, before excluding generation-error episodes. These counts describe the evaluation budget, rather than the number of episodes.

\begin{table}[t]
\centering
\caption{\textbf{Evaluation episodes per setting.} Counts are reported for one target model, method, and setting before error exclusions. Each benchmark follows its official evaluation protocol.}
\label{tab:supp_episodes}
\tabcapbelow
\setlength{\tabcolsep}{8pt}
\scriptsize
\resizebox{\linewidth}{!}{%
\begin{tabular}{@{}lp{0.47\linewidth}r@{}}
\toprule
Benchmark & Evaluation configuration & Episodes \\
\midrule
LIBERO~\cite{libero} & 4 suites $\times$ 10 tasks $\times$ 50 initial states & 2,000 \\
LIBERO-Plus~\cite{libero_plus} & One run per variation case & 10,030 \\
LIBERO-Pro~\cite{libero_pro} & 4 suites $\times$ 5 variants $\times$ 10 tasks $\times$ 50 initial states & 10,000 \\
SimplerEnv~\cite{simplerenv} & WidowX: 4 tasks $\times$ 24 episodes & 96 \\
\bottomrule
\end{tabular}%
}
\vspace{-15pt}
\end{table}

\subpara{LIBERO and its variants.}
For LIBERO-Plus, we execute every variation case once, giving 2,591, 2,402, 2,518, and 2,519 episodes for Goal, Spatial, Object, and Long, respectively. For LIBERO-Pro, we evaluate the five variants env, lan, object, swap, and task. Each variant contains ten tasks and 50 initial states per task in each suite, giving 2,500 episodes per suite across the five variants.

\subpara{SimplerEnv.}
We evaluate StarVLA-FAST on the WidowX embodiment with four tasks: placing a spoon on a towel, placing a carrot on a plate, stacking a green cube on a yellow cube, and placing an eggplant in a basket.

\subpara{Reporting.}
Generation errors are counted as failures in success rates but excluded from efficiency metrics; details are provided in \cref{sec:supp_errors}.

\subsection{Training Data}
\label{sec:supp_training}
\cref{tab:supp_training} reports the draft training set sizes on LIBERO. For each target, baselines use 27,548 samples collected through rollouts without action noise, and VSpark uses the same number collected through rollouts with calibrated action noise. VSpark is trained only on the latter samples; we do not combine the two sets.

\begin{table}[t]
\centering
\caption{\textbf{Draft training set sizes.} Baselines use rollouts without action noise; VSpark uses rollouts with calibrated action noise. Sample counts and optimizer updates are matched.}
\label{tab:supp_training}
\vspace{-0.2em}
\setlength{\tabcolsep}{13pt}
\scriptsize
\resizebox{\linewidth}{!}{%
\begin{tabular}{@{}lrr@{}}
\toprule
Target & Baselines (no noise) & VSpark (noise) \\
\midrule
$\pi_0$-FAST & 27,548 & 27,548 \\
StarVLA-FAST & 27,548 & 27,548 \\
ActionCodec & 27,548 & 27,548 \\
G0.5 & 27,548 & 27,548 \\
\bottomrule
\end{tabular}%
}
\vspace{-15pt}
\end{table}
\begin{table*}[t]
\centering
\caption{\textbf{Model checkpoints and baseline repositories.} Identifiers link to Hugging Face for target models and GitHub for baselines. LIBERO checkpoints are shared across LIBERO, LIBERO-Plus, and LIBERO-Pro; the WidowX checkpoint is used for SimplerEnv.}
\label{tab:supp_model_resources}
\vspace{-10pt}
\setlength{\tabcolsep}{40pt}
\scriptsize
\resizebox{\linewidth}{!}{%
\begin{tabular}{ll}
\toprule
Model / method & Repository / checkpoint \\
\midrule
\multicolumn{2}{c}{\textit{Target models}} \\
\midrule
$\pi_0$-FAST~\cite{fast} & \href{https://huggingface.co/lerobot/pi0fast-libero}{\texttt{lerobot/pi0fast-libero}} \\
StarVLA-FAST (LIBERO)~\cite{community2026starvla} & \href{https://huggingface.co/StarVLA/Qwen2.5-VL-FAST-LIBERO-4in1}{\texttt{StarVLA/Qwen2.5-VL-FAST-LIBERO-4in1}} \\
& \texttt{checkpoints/steps\_30000\_pytorch\_model.pt} \\
StarVLA-FAST (WidowX)~\cite{community2026starvla} & \href{https://huggingface.co/StarVLA/Qwen-FAST-Bridge-RT-1}{\texttt{StarVLA/Qwen-FAST-Bridge-RT-1}} \\
& \texttt{checkpoints/steps\_10000\_pytorch\_model.pt} \\
ActionCodec~\cite{actioncodec} & \href{https://huggingface.co/ZibinDong/SmolVLM2-2.2B-ActionCodec-AR-LIBERO}{\texttt{ZibinDong/SmolVLM2-2.2B-ActionCodec-AR-LIBERO}} \\
G0.5~\cite{g05} & \href{https://huggingface.co/ZibinDong/g05-libero-ar}{\texttt{ZibinDong/g05-libero-ar}} \\
\midrule
\multicolumn{2}{c}{\textit{Speculative decoding baselines}} \\
\midrule
EAGLE-3~\cite{eagle3} & \href{https://github.com/SafeAILab/EAGLE}{\texttt{SafeAILab/EAGLE}} \\
P-EAGLE~\cite{peagle} & \href{https://github.com/vllm-project/speculators}{\texttt{vllm-project/speculators}} \\
DFlash~\cite{dflash} & \href{https://github.com/z-lab/dflash}{\texttt{z-lab/dflash}} \\
DFlash2~\cite{dflash2} & \href{https://github.com/vllm-project/speculators}{\texttt{vllm-project/speculators}} \\
DSpark~\cite{dspark} & \href{https://github.com/deepseek-ai/DeepSpec}{\texttt{deepseek-ai/DeepSpec}} \\
\bottomrule
\end{tabular}%
}
\vspace{-10pt}
\end{table*}

The training schedule follows \cref{sec:experiments}; the comparison therefore changes the training states while keeping the sample count and optimizer update budget fixed. For SimplerEnv (WidowX), we train both the baseline drafts and the VSpark draft for StarVLA-FAST on 26,413 rollouts from BridgeData V2.

\subsection{Model and Baseline Resources}
\label{sec:supp_resources}
\cref{tab:supp_model_resources} lists the target checkpoints and baseline repositories. We use the same LIBERO checkpoint for LIBERO, LIBERO-Plus, and LIBERO-Pro, and a separate StarVLA-FAST checkpoint for SimplerEnv WidowX. The StarVLA-FAST checkpoint files are specified explicitly.

\subsection{Draft Architecture}
\label{sec:supp_draft_architecture}
\begin{table}[t]
\centering
\caption{\textbf{Draft specifications.}
Target sizes follow the public model labels; draft sizes are rounded to the nearest million parameters.
KV denotes key and value heads; FFN dim. is the intermediate dim.}
\label{tab:supp_draft_architecture}
\vspace{-6pt}
\setlength{\tabcolsep}{4pt}
\scriptsize
\resizebox{\linewidth}{!}{%
\begin{tabular}{lrrrr}
\toprule
Target & $\pi_0$-FAST & StarVLA-FAST & ActionCodec & G0.5 \\
\midrule
Target size & 3B & 3B & 2.2B & 3B \\
VSpark Draft size & 205M & 147M & 119M & 150M \\
\midrule
Layers & 1 & 1 & 1 & 1 \\
Hidden dim. & 2,048 & 2,048 & 2,048 & 2,048 \\
Query heads & 8 & 16 & 32 & 8 \\
KV heads & 1 & 2 & 32 & 2 \\
Head dim. & 256 & 128 & 64 & 256 \\
FFN dim. & 16,384 & 11,008 & 8,192 & 6,144 \\
RoPE $\theta$ & $10^4$ & $10^6$ & $1.3\times10^5$ & $10^7$ \\
\midrule
Output vocabulary & 1,030 & 2,050 & 2,056 & 4,107 \\
\bottomrule
\end{tabular}%
}
\vspace{-20pt}
\end{table}

\cref{tab:supp_draft_architecture} summarizes the draft settings shared by VSpark and all baselines for each target. Each draft has one layer by default; the five-layer experiments extend the same architecture by increasing only the layer count. Its hidden and head dimensions, attention head counts, FFN intermediate dimension, and RoPE base follow the corresponding target configuration. The output head is restricted to the action vocabulary, whose size is determined by the action token count of each tokenizer.

\subsection{Feature Tap Selection}
\label{sec:supp_features}
\cref{tab:supp_taps} lists the target layers used for feature conditioning. Indices start at one and refer to the outputs of Transformer blocks. For each target, we use five layers distributed approximately uniformly from shallow to deep blocks.

\begin{table}[t]
\centering
\caption{\textbf{Target layers for feature conditioning.} Indices refer to Transformer block outputs and start at one. Each target supplies five layers spanning shallow to deep representations.}
\label{tab:supp_taps}
\tabcapbelow
\setlength{\tabcolsep}{30pt}
\scriptsize
\resizebox{\linewidth}{!}{%
\begin{tabular}{lc}
\toprule
Target & Feature layers \\
\midrule
$\pi_0$-FAST & 2, 5, 9, 13, 16 \\
StarVLA-FAST & 2, 10, 18, 26, 34 \\
ActionCodec & 2, 7, 12, 17, 22 \\
G0.5 & 2, 7, 12, 17, 22 \\
\bottomrule
\end{tabular}%
}
\tabbelow
\end{table}
\begin{table}[t]
\centering
\caption{\textbf{Calibrated action noise on LIBERO.} We report the marginal standard deviations of the fitted zero-mean Gaussian, converted to action-command perturbations in millimeters and degrees. Each model uses 64 calibration inputs, with 16 from each suite.}
\label{tab:supp_noise}
\tabcapbelow
\setlength{\tabcolsep}{6.5pt}
\scriptsize
\resizebox{\linewidth}{!}{%
\begin{tabular}{lrrrrrr}
\toprule
& \multicolumn{3}{c}{Translation (mm)} & \multicolumn{3}{c}{Rotation ($^\circ$)} \\
\cmidrule(lr){2-4}\cmidrule(lr){5-7}
Target & $\sigma_x$ & $\sigma_y$ & $\sigma_z$ & $\sigma_{r_x}$ & $\sigma_{r_y}$ & $\sigma_{r_z}$ \\
\midrule
$\pi_0$-FAST & 3.88 & 7.10 & 3.08 & 0.24 & 0.66 & 0.90 \\
StarVLA-FAST & 4.88 & 6.49 & 6.17 & 0.42 & 1.36 & 1.67 \\
ActionCodec & 2.94 & 2.73 & 3.31 & 0.33 & 0.49 & 0.34 \\
G0.5 & 1.17 & 0.87 & 2.30 & 0.22 & 0.39 & 0.38 \\
\bottomrule
\end{tabular}%
}
\vspace{-15pt}
\end{table}

\subsection{Calibrating Action Noise}
\label{sec:supp_noise}
\subpara{Calibration samples.}
For each target, we use an initial one-layer draft trained on the original training data. Calibration uses 16 training samples from each of the four LIBERO suites, giving 64 input samples per target. For each target token sequence, we choose up to four uniformly spaced starting positions. At each position, we replace a block of seven consecutive tokens with the draft's top-1 path, keeping the preceding target prefix and the remaining target suffix fixed. We decode the original and modified sequences into actions and collect their differences at each timestep, retaining valid decodings.

\subpara{Noise statistics.}
The residual count in \cref{eq:noise_covariance} includes all valid timestep residuals from the 64 calibration inputs, not the number of input samples. We use the empirical second moment without subtracting the residual mean and retain all off-diagonal entries of the $6\times6$ matrix. \cref{tab:supp_noise} reports the marginal standard deviations for each target after conversion to action-command perturbations in millimeters for translation and degrees for rotation. The scales differ across targets and action dimensions, so calibration does not reduce to a single shared scalar.

\section{Further Analysis}
\label{sec:supp_analysis}
Unless otherwise specified, the analyses below use LIBERO and the settings in \cref{sec:experiments}. We examine the components of VSpark, the resulting acceptance lengths, and the limits of alternatives that change the target's tokens.

\subsection{Feature Conditioning and Regression}
\begin{table}[t]
\centering
\caption{\textbf{Feature alignment component ablation} on $\pi_0$-FAST.
We independently enable feature conditioning and feature regression.
All settings use one-layer drafts, training only on noise rollouts,
continuation reuse, and early stopping.
Results are averaged over LIBERO suites.}
\label{tab:feature_alignment_ablation}
\vspace{-6pt}
\setlength{\tabcolsep}{13pt}
\scriptsize
\resizebox{\linewidth}{!}{%
\begin{tabular}{cccc}
\toprule
Feature input & Regression
& $\tau\uparrow$ & Speedup$\uparrow$ \\
\midrule
$\times$ & $\times$ & 3.23 & $1.68\times$ \\
$\checkmark$  & $\times$ & 3.27 & $1.70\times$ \\
$\times$ & $\checkmark$  & 3.67 & $1.78\times$ \\
\rowcolor{aliceblue}
$\checkmark$ & $\checkmark$ & \textbf{3.94} & $\mathbf{1.93}\times$ \\
\bottomrule
\end{tabular}%
}
\vspace{-6pt}
\end{table}
\cref{tab:feature_alignment_ablation} separates the two parts of feature alignment on $\pi_0$-FAST: conditioning the draft on target features and regressing the target feature before the output head. All variants use the same noise rollout training, continuation reuse, and early stopping. Conditioning alone changes $\tau$ only slightly, from 3.23 to 3.27, whereas feature regression alone raises it to 3.67. Combining both gives $\tau=3.94$ and a speedup of $1.93\times$, compared with $1.68\times$ when both are disabled. Feature regression provides the larger individual gain, while conditioning improves the result when paired with regression. Thus, the benefit of feature alignment is not explained by supplying additional target features alone.

\begin{table}[t]
\centering
\caption{\textbf{Continuation reuse component ablation} on $\pi_0$-FAST.
We independently enable continuation reuse and early stopping.
All settings use one-layer drafts, training only on noise rollouts,
and feature alignment.
Results are averaged over LIBERO suites.
}
\label{tab:continuation_reuse_ablation}
\vspace{-6pt}
\setlength{\tabcolsep}{16pt}
\scriptsize
\resizebox{\linewidth}{!}{%
\begin{tabular}{cccc}
\toprule
Reuse & Early stop
& $\tau\uparrow$ & Speedup$\uparrow$ \\
\midrule
$\times$ & $\times$ & 3.87 & $1.84\times$ \\
$\checkmark$  & $\times$ & \textbf{3.94} & $1.86\times$ \\
$\times$ & $\checkmark$  & 3.87 & $1.86\times$ \\
\rowcolor{aliceblue}
$\checkmark$ & $\checkmark$ & \textbf{3.94} & $\mathbf{1.93}\times$ \\
\bottomrule
\end{tabular}%
}
\vspace{-6pt}
\end{table}
\subsection{Continuation Reuse and Early Stopping}
\cref{tab:continuation_reuse_ablation} independently enables continuation reuse and early stopping on $\pi_0$-FAST while keeping feature alignment and noise rollout training fixed. Reuse raises $\tau$ from 3.87 to 3.94 with or without early stopping. Early stopping leaves $\tau$ unchanged at the reported precision but improves speedup, from $1.84\times$ to $1.86\times$ without reuse and from $1.86\times$ to $1.93\times$ with reuse. The two components therefore contribute in different ways: reuse supplies candidates that extend the accepted path, while early stopping reduces the work spent constructing the tree.

\begin{table}[t]
\centering
\caption{\textbf{Statistics of accepted draft length.}
We report the percentage of verification cycles accepting
at least $k$ consecutive draft tokens.
Mean committed length $\tau$ additionally includes target
correction or additional tokens.
Results are averaged over LIBERO suites.
}
\label{tab:accepted_length}
\vspace{-6pt}
\setlength{\tabcolsep}{4pt}
\scriptsize
\resizebox{\linewidth}{!}{%
\begin{tabular}{l*{8}{c}}
\toprule
Method
& $\geq 1$ & $\geq 2$ & $\geq 3$ & $\geq 4$
& $\geq 5$ & $\geq 6$ & $\geq 7$ & $\tau\uparrow$ \\
\midrule
\multicolumn{9}{c}{\textit{$\pi_0$-FAST}} \\
\midrule
DSpark
& 91.63 & 79.42 & 32.00 & 10.43
& 1.87 & 0.54 & 0.25 & 3.16 \\
\rowcolor{aliceblue}
\textbf{VSpark}
& \textbf{95.19} & \textbf{81.26}
& \textbf{54.76} & \textbf{33.94}
& \textbf{16.71} & \textbf{7.91}
& \textbf{3.96} & \textbf{3.94} \\
\midrule
\multicolumn{9}{c}{\textit{ActionCodec}} \\
\midrule
DSpark
& 83.34 & 66.89 & 47.30 & 31.25
& 20.82 & 12.32 & 6.97 & 3.69 \\
\rowcolor{aliceblue}
\textbf{VSpark}
& \textbf{92.80} & \textbf{77.98}
& \textbf{58.29} & \textbf{39.83}
& \textbf{26.83} & \textbf{15.93}
& \textbf{9.31} & \textbf{4.21} \\
\bottomrule
\end{tabular}%
}
\vspace{-6pt}
\end{table}
\subsection{Distribution of Accepted Draft Lengths}
\cref{tab:accepted_length} reports the fraction of verification cycles accepting at least $k$ consecutive draft tokens. These counts exclude the correction or additional token predicted by the target, which is included in $\tau$. On $\pi_0$-FAST, VSpark increases the fraction of cycles accepting at least three draft tokens from 32.00\% to 54.76\%, and at least five from 1.87\% to 16.71\%. On ActionCodec, the corresponding fractions rise from 47.30\% to 58.29\% and from 20.82\% to 26.83\%. Improvements hold at every reported length for both targets. The gain in mean committed length thus reflects more frequent long accepted paths, rather than only a greater chance of accepting the first draft token.

\begin{table}[t]
\centering
\caption{\textbf{Nearest-embedding token replacement without drafting.}
Each eligible action token is replaced with probability $p$
by its nearest distinct token in the action vocabulary
under cosine distance.
Online replacement conditions subsequent generation on the
modified prefix; post-hoc replacement modifies the completed
AR chunk before action decoding and execution.
We report success rates (\%) on LIBERO Goal.}
\label{tab:embedding_replacement}
\vspace{-6pt}
\setlength{\tabcolsep}{6pt}
\scriptsize
\resizebox{\linewidth}{!}{%
\begin{tabular}{c*{4}{c}}
\toprule
& \multicolumn{2}{c}{$\pi_0$-FAST}
& \multicolumn{2}{c}{ActionCodec} \\
\cmidrule(lr){2-3}\cmidrule(lr){4-5}
Replacement probability
& Online & Post-hoc
& Online & Post-hoc \\
\midrule
0\% (AR) & 90.6 & 90.6 & 85.4 & 85.4 \\
\midrule
1\% & 83.2 & 79.4 & 86.4 & 87.0 \\
5\% & 59.4 & 52.0 & 79.2 & 81.0 \\
10\%  & 35.0 & 30.6 & 68.0 & 69.8 \\
25\%  & 7.6  & 9.4  & 29.0 & 29.2 \\
50\%  & 2.0  & 1.4  & 2.2  & 3.2 \\
100\% & 0.0  & 0.0  & 0.0  & 0.0 \\
\bottomrule
\end{tabular}%
}
\vspace{-6pt}
\end{table}

\begin{table*}[t]
\centering
\captionsetup{width=\textwidth}
\caption{\textbf{Applying VSpark to other draft architectures on LIBERO.}
We train each baseline on rollouts with calibrated action noise and add
feature alignment, continuation reuse, and early stopping.
Speedup uses the same target autoregressive reference for each baseline
and its VSpark variant. Avg. is the arithmetic mean over LIBERO suites.}
\label{tab:vspark_transfer}
\vspace{-6pt}
\setlength{\tabcolsep}{9.7pt}
\scriptsize
\resizebox{\linewidth}{!}{%
\begin{tabular}{l*{10}{c}}
\toprule
& \multicolumn{2}{c}{Goal}
& \multicolumn{2}{c}{Spatial}
& \multicolumn{2}{c}{Object}
& \multicolumn{2}{c}{Long}
& \multicolumn{2}{c}{Avg.} \\
\cmidrule(lr){2-3}\cmidrule(lr){4-5}
\cmidrule(lr){6-7}\cmidrule(lr){8-9}
\cmidrule(lr){10-11}
Method
& $\tau\uparrow$ & Speedup$\uparrow$
& $\tau\uparrow$ & Speedup$\uparrow$
& $\tau\uparrow$ & Speedup$\uparrow$
& $\tau\uparrow$ & Speedup$\uparrow$
& $\tau\uparrow$ & Speedup$\uparrow$ \\
\midrule
\multicolumn{11}{c}{\textit{$\pi_0$-FAST}} \\
\midrule
EAGLE3
& 3.09 & $1.22\times$ & 3.12 & $1.24\times$
& 3.31 & $1.31\times$ & 3.17 & $1.27\times$
& 3.17 & $1.26\times$ \\
\rowcolor{aliceblue}
+ VSpark
& \textbf{4.31} & $\mathbf{1.60}\times$
& \textbf{4.48} & $\mathbf{1.65}\times$
& \textbf{4.59} & $\mathbf{1.71}\times$
& \textbf{4.34} & $\mathbf{1.57}\times$
& \textbf{4.43} & $\mathbf{1.63}\times$ \\
\midrule
P-EAGLE
& 2.94 & $1.45\times$ & 2.97 & $1.49\times$
& 3.12 & $1.57\times$ & 3.00 & $1.48\times$
& 3.01 & $1.50\times$ \\
\rowcolor{aliceblue}
+ VSpark
& \textbf{3.65} & $\mathbf{1.74}\times$
& \textbf{3.78} & $\mathbf{1.78}\times$
& \textbf{3.82} & $\mathbf{1.79}\times$
& \textbf{3.64} & $\mathbf{1.70}\times$
& \textbf{3.72} & $\mathbf{1.76}\times$ \\
\midrule
DFlash
& 2.33 & $1.39\times$ & 2.42 & $1.46\times$
& 2.44 & $1.46\times$ & 2.30 & $1.36\times$
& 2.37 & $1.41\times$ \\
\rowcolor{aliceblue}
+ VSpark
& \textbf{2.94} & $\mathbf{1.60}\times$
& \textbf{3.05} & $\mathbf{1.71}\times$
& \textbf{3.04} & $\mathbf{1.68}\times$
& \textbf{2.88} & $\mathbf{1.57}\times$
& \textbf{2.98} & $\mathbf{1.64}\times$ \\

\midrule
\multicolumn{11}{c}{\textit{ActionCodec}} \\
\midrule
EAGLE3
& 3.92 & $1.58\times$ & 3.76 & $1.57\times$
& 4.26 & $1.73\times$ & 4.01 & $1.65\times$
& 3.99 & $1.63\times$ \\
\rowcolor{aliceblue}
+ VSpark
& \textbf{4.08} & $\mathbf{1.63}\times$
& \textbf{3.85} & $\mathbf{1.62}\times$
& \textbf{4.40} & $\mathbf{1.76}\times$
& \textbf{4.15} & $1.65\times$
& \textbf{4.12} & $\mathbf{1.66}\times$ \\
\midrule
P-EAGLE
& 3.73 & $1.88\times$ & 3.66 & $1.90\times$
& 4.09 & $2.07\times$ & 3.83 & $1.96\times$
& 3.83 & $1.95\times$ \\
\rowcolor{aliceblue}
+ VSpark
& \textbf{4.15} & $\mathbf{2.03}\times$
& \textbf{3.96} & $\mathbf{2.03}\times$
& \textbf{4.44} & $\mathbf{2.19}\times$
& \textbf{4.19} & $\mathbf{2.06}\times$
& \textbf{4.19} & $\mathbf{2.08}\times$ \\
\midrule
DFlash
& 3.30 & $1.84\times$ & 3.16 & $1.83\times$
& 3.64 & $2.02\times$ & 3.35 & $1.90\times$
& 3.36 & $1.90\times$ \\
\rowcolor{aliceblue}
+ VSpark
& \textbf{3.94} & $\mathbf{2.09}\times$
& \textbf{3.83} & $\mathbf{2.08}\times$
& \textbf{4.36} & $\mathbf{2.30}\times$
& \textbf{4.12} & $\mathbf{2.18}\times$
& \textbf{4.06} & $\mathbf{2.16}\times$ \\
\bottomrule
\end{tabular}%
}
\vspace{-6pt}
\end{table*}
\subsection{Nearest-Embedding Token Replacement}
\label{sec:supp_replacement}
The relaxed verification in \cref{tab:embedding_relaxed} only considers tokens proposed by the draft. This restriction may help preserve success even when embedding distance does not reliably reflect action discrepancy. Here, we test whether embedding proximity alone is sufficient by replacing tokens without a draft model. At each eligible action token, we replace the target's token with probability $p$ by its nearest distinct action token under cosine distance. Online replacement conditions subsequent generation on the modified prefix. Post-hoc replacement instead changes the completed target sequence before action decoding and execution. Both settings are evaluated on all 500 LIBERO Goal episodes, including generation failures.

As shown in \cref{tab:embedding_replacement}, even a 1\% replacement probability lowers $\pi_0$-FAST success from 90.6\% to 83.2\% online and 79.4\% post-hoc. At 5\%, success falls further to 59.4\% and 52.0\%, respectively. ActionCodec is less sensitive at low replacement probabilities: its online and post-hoc success rates are 86.4\% and 87.0\% at 1\%, and fall to 79.2\% and 81.0\% at 5\%. At 10\%, success drops to 35.0\% and 30.6\% for $\pi_0$-FAST and to 68.0\% and 69.8\% for ActionCodec. At 50\% replacement, success is at most 3.2\% for either model, and replacing every eligible token gives zero success. This degradation is consistent with draft proposals helping preserve success under relaxed verification, although the two experiments use different replacement rules and evaluation scopes. Without this restriction, nearby token embeddings can decode to substantially different actions, and success degrades even when subsequent tokens are generated under the modified prefix. These results further support keeping verification exact.

\subsection{Applying VSpark to Other Draft Architectures}
We examine whether the improvements extend beyond the DSpark draft used in VSpark. \cref{tab:vspark_transfer} applies noise rollout training, feature alignment, continuation reuse, and early stopping to EAGLE-3, P-EAGLE, and DFlash. Each baseline and its augmented variant use the same target autoregressive reference for speedup.

On $\pi_0$-FAST, average speedup rises from $1.26\times$ to $1.63\times$ for EAGLE-3, from $1.50\times$ to $1.76\times$ for P-EAGLE, and from $1.41\times$ to $1.64\times$ for DFlash. All three also improve on ActionCodec, reaching $1.66\times$, $2.08\times$, and $2.16\times$, respectively. Acceptance length increases in every model--suite comparison, although the speedup gain varies with the draft architecture. In particular, EAGLE-3 reaches $\tau=4.43$ on $\pi_0$-FAST but a speedup of $1.63\times$, illustrating again that acceptance length alone does not determine decoding time. Across the three draft architectures and two targets evaluated here, the combined design choices improve both average acceptance length and speedup.

\begin{table}[t]
\centering
\caption{\textbf{Kalman prediction errors versus draft action discrepancies.}
We compare Kalman prediction errors with action differences
induced by replacing a single rejected token while keeping
the remaining target tokens fixed.
Results are reported separately for translation and rotation,
at the first timestep and across the full action chunk.
Ratio denotes the median of $e_K/e_D$, where $e_K$ is the
Kalman prediction error and $e_D$ is the action discrepancy
caused by token replacement.
$P(e_K > e_D)$ denotes the fraction of cases where
the Kalman error exceeds the draft discrepancy.}
\label{tab:kalman_analysis}
\tabcapbelow
\setlength{\tabcolsep}{2.5pt}
\scriptsize
\resizebox{\linewidth}{!}{%
\begin{tabular}{llcccc}
\toprule
& & \multicolumn{2}{c}{$\pi_0$-FAST}
& \multicolumn{2}{c}{ActionCodec} \\
\cmidrule(lr){3-4}\cmidrule(lr){5-6}
Component & Scope
& Ratio & $P(e_K > e_D)$
& Ratio & $P(e_K > e_D)$ \\
\midrule
Translation & First step
& 2.68$\times$ & 85.00\%
& 17.76$\times$ & 97.04\% \\
            & Full chunk
& 4.56$\times$ & 94.69\%
& 15.28$\times$ & 99.61\% \\
\midrule
Rotation & First step
& 3.03$\times$ & 87.44\%
& 18.98$\times$ & 96.38\% \\
         & Full chunk
& 4.66$\times$ & 96.30\%
& 10.11$\times$ & 99.45\% \\
\bottomrule
\end{tabular}%
}
\tabbelow
\end{table}
\begin{table*}[t]
\centering
\caption{\textbf{Numerical precision and agreement with autoregressive decoding on LIBERO.}
All values are percentages. BF16 reports mean SR with 95\% Student-$t$ confidence intervals. Agreement is the fraction of paired episodes with the same success/failure outcome under AR and VSpark. Avg. averages the four suites within each seed before computing the mean and confidence interval.}
\label{tab:supp_numerical_precision}
\tabcapbelow
\setlength{\tabcolsep}{13pt}
\scriptsize
\resizebox{\linewidth}{!}{%
\begin{tabular}{lcccccc}
\toprule
& \multicolumn{3}{c}{BF16} & \multicolumn{3}{c}{FP32} \\
\cmidrule(lr){2-4}\cmidrule(lr){5-7}
Suite & AR SR [95\% CI] & VSpark SR [95\% CI] & Agreement & AR SR & VSpark SR & Agreement \\
\midrule
\multicolumn{7}{c}{\textit{$\pi_0$-FAST}} \\
\midrule
Goal & 91.64 [89.48, 93.80] & 91.40 [90.16, 92.64] & 88.56 & 91.00 & 91.00 & 100.00 \\
Spatial & 94.92 [93.89, 95.95] & 95.48 [94.37, 96.59] & 94.80 & 95.80 & 95.80 & 100.00 \\
Object & 97.40 [97.15, 97.65] & 96.68 [96.35, 97.01] & 95.68 & 96.40 & 96.40 & 100.00 \\
Long & 86.87 [86.11, 87.63] & 85.33 [82.34, 88.33] & 82.47 & 85.20 & 85.20 & 100.00 \\
\rowcolor{aliceblue}
Avg. & 92.71 [92.17, 93.24] & 92.22 [91.64, 92.80] & 90.38 & 92.10 & 92.10 & 100.00 \\
\midrule
\multicolumn{7}{c}{\textit{ActionCodec}} \\
\midrule
Goal & 88.12 [85.98, 90.26] & 88.16 [86.54, 89.78] & 85.48 & 80.80 & 80.80 & 100.00 \\
Spatial & 83.96 [82.89, 85.03] & 84.16 [81.03, 87.29] & 81.16 & 79.80 & 79.80 & 100.00 \\
Object & 93.24 [93.13, 93.35] & 92.76 [92.44, 93.08] & 93.04 & 93.40 & 93.40 & 100.00 \\
Long & 77.88 [76.80, 78.96] & 79.36 [77.01, 81.71] & 74.76 & 75.00 & 75.00 & 100.00 \\
\rowcolor{aliceblue}
Avg. & 85.80 [85.06, 86.54] & 86.11 [84.68, 87.54] & 83.61 & 82.25 & 82.25 & 100.00 \\
\bottomrule
\end{tabular}%
}
\tabbelow
\end{table*}
\subsection{Precision Limits of Kalman Prediction}
We examine whether Kalman filtering can predict the fine action differences associated with rejected draft tokens. \cref{tab:kalman_analysis} compares Kalman prediction errors with action discrepancies caused by replacing a single rejected token while keeping the remaining target tokens fixed. We report translation and rotation separately, both at the first timestep and across the full action chunk.

For $\pi_0$-FAST, the median ratio of Kalman error to draft action discrepancy ranges from 2.68 to 4.66; for ActionCodec, it ranges from 10.11 to 18.98. Kalman errors exceed draft discrepancies in 85.00--96.30\% and 96.38--99.61\% of cases, respectively. Under this evaluation, the prediction error is typically larger than the action difference induced by a token mismatch. These results suggest that the evaluated Kalman predictor lacks the precision needed to resolve these fine differences.

\subsection{Lossless Verification}
\label{sec:supp_precision}
\cref{tab:supp_numerical_precision} compares AR and VSpark on $\pi_0$-FAST and ActionCodec under BF16 and FP32. For BF16, we report mean success rates and 95\% Student-$t$ confidence intervals over five seeds (0--4). All other LIBERO tables use simulator seed 7, for which repeated AR runs produced identical results. Agreement measures whether paired episodes have the same success/failure outcome. The suite-averaged SR is 92.71\% for AR and 92.22\% for VSpark on $\pi_0$-FAST, and 85.80\% and 86.11\% on ActionCodec. These correspond to differences of $-0.49$ and $+0.31$ percentage points, respectively, indicating similar aggregate task performance.

\subpara{Exact verification and finite precision.}
VSpark retains exact token verification: under exact arithmetic, greedy verification preserves the target's autoregressive output, following the lossless principle of speculative decoding~\cite{leviathan2023,chen2023}. In finite precision, parallel tree verification and sequential AR decoding can evaluate the same prefix with different floating-point operation orders. Small numerical differences can change the selected token when leading logits are close, after which subsequent actions and observations may diverge. Such numerical discrepancies are also documented for speculative decoding in LLMs~\cite{chen2023,vllm_speculative_numerics}.

\subpara{FP32 comparison.}
With computation cast to FP32, AR and VSpark have identical success rates and 100\% episode-outcome agreement in every suite for both targets at seed 0. We additionally confirm that their generated action tokens match exactly in these runs. This supports finite-precision effects during verification as the source of the BF16 discrepancies, rather than a relaxation of the acceptance rule. The FP32 result is an empirical observation for these targets and this seed; the theoretical lossless property concerns exact verification under matching target computations.

\begin{table}[t]
\centering
\caption{\textbf{Generation-error rates across evaluation methods.}
For each method, we report the fraction of episodes containing generation errors
in its own rollouts, including failures to decode generated tokens into valid actions.
Values are averaged over target models and benchmark suites.
Task failures and token-limit hits alone are not counted
as generation errors.}
\label{tab:generation_errors}
\vspace{-6pt}
\setlength{\tabcolsep}{7.8pt}
\scriptsize
\resizebox{\linewidth}{!}{%
\begin{tabular}{lccc}
\toprule
Method & LIBERO & LIBERO-Plus & LIBERO-Pro \\
\midrule
Target AR & 0.35\% & 1.65\% & 1.14\% \\
\midrule
EAGLE3     & 0.33\% & 1.62\% & 1.18\% \\
P-EAGLE    & 0.45\% & 1.63\% & 1.10\% \\
DFlash     & 0.38\% & 1.63\% & 1.04\% \\
DFlash2    & 0.48\% & 1.64\% & 1.16\% \\
DSpark     & 0.38\% & 1.55\% & 1.20\% \\
\rowcolor{aliceblue}
VSpark     & 0.28\% & 1.62\% & 1.17\% \\
\bottomrule
\end{tabular}%
}
\vspace{-6pt}
\end{table}
\subsection{Generation Errors}
\label{sec:supp_errors}
Generation errors include cases where the generated token sequence cannot be decoded into valid actions, such as when the target model generates text tokens instead of action tokens. We exclude the entire affected episode from acceptance-length and timing statistics. A token-limit chunk reaches the maximum generation length before an end-of-sequence token is produced, for example when the model repeatedly generates the same token; only that chunk is excluded from these statistics. Task failures and token-limit hits alone are not classified as generation errors. \cref{tab:generation_errors} reports the fraction of episodes containing generation errors in each method's own rollouts before exclusion, averaged over four suites and four target VLAs.

Error rates are low across all methods: 0.28--0.48\% on LIBERO, 1.55--1.65\% on LIBERO-Plus, and 1.04--1.20\% on LIBERO-Pro. VSpark has rates of 0.28\%, 1.62\%, and 1.17\%, respectively, compared with 0.35\%, 1.65\%, and 1.14\% for target autoregressive decoding. The observed error rates are therefore comparable in magnitude. These aggregate rollout statistics describe the frequency of episode exclusions from acceptance-length and timing statistics due to generation errors.

\begin{table}[t]
\centering
    \caption{\textbf{Time breakdown} of $\pi_0$-FAST on LIBERO. Times are ms per output token, averaged over four suites. Draft includes tree construction; Target denotes AR decoding or speculative verification; Action denotes action decoding.}
\label{tab:supp_latency_breakdown}
\tabcapbelow
\setlength{\tabcolsep}{6pt}
\scriptsize
\resizebox{\linewidth}{!}{%
\begin{tabular}{lrrrrrr}
\toprule
Method & Draft & Target & Prefill & Action & Other & Total \\
\midrule
AR & - & 12.02 & 2.40 & 0.02 & 0.03 & 14.47 \\
DSpark & 1.58 & 4.23 & 2.41 & 0.02 & 0.70 & 8.93 \\
\rowcolor{aliceblue}
VSpark & 1.32 & 3.30 & 2.24 & 0.02 & 0.56 & 7.44 \\
\bottomrule
\end{tabular}%
}
\tabbelow
\end{table}
\subsection{Time Breakdown}
\label{sec:supp_latency}
\cref{tab:supp_latency_breakdown} separates the measured time on $\pi_0$-FAST into drafting and tree construction, target decoding or verification, prefill, action decoding, and other processing. We use CUDA events recorded every tenth action chunk in the LIBERO evaluation logs. Within each suite, we divide each component's accumulated time by the total number of output tokens in the same profiling samples, then average the four suite values equally.

Compared with DSpark, VSpark reduces total time per output token from 8.93 to 7.44 ms, a reduction of 16.7\%. Target verification time falls from 4.23 to 3.30 ms, while drafting and tree construction decrease from 1.58 to 1.32 ms. The reduction in verification time is the largest contribution to the total saving. On this target, the higher acceptance length is accompanied by lower drafting and tree construction time per output token.

\subsection{Removing Individual Components}
\label{sec:supp_component_removal}
\begin{table}[t]
\centering
\caption{\textbf{Independent component removal} on $\pi_0$-FAST. Each variant removes one component from VSpark. Continuation reuse includes early stopping. Results are averaged over four LIBERO suites.}
\label{tab:supp_component_removal}
\tabcapbelow
\setlength{\tabcolsep}{15pt}
\scriptsize
\resizebox{\linewidth}{!}{%
\begin{tabular}{lrr}
\toprule
Configuration & $\tau\uparrow$ & Speedup$\uparrow$ \\
\midrule
\rowcolor{aliceblue}
VSpark & \textbf{3.94} & $\mathbf{1.93}\times$ \\
\quad w/o Action Noise & 3.87 & $1.90\times$ \\
\quad w/o Feature Alignment & 3.23 & $1.68\times$ \\
\quad w/o Continuation Reuse & 3.87 & $1.84\times$ \\
\midrule
DSpark & 3.16 & $1.60\times$ \\
\bottomrule
\end{tabular}%
}
\tabbelow
\end{table}
\cref{tab:supp_component_removal} removes each component independently from the complete VSpark configuration on $\pi_0$-FAST. Acceptance length and speedup are averaged over the four LIBERO suites. As in \cref{tab:component_ablation}, continuation reuse includes early stopping; removing this component disables both. This complements the cumulative additions by measuring each component's contribution when the others remain enabled.

On $\pi_0$-FAST, removing feature alignment gives the largest drop in acceptance length, from 3.94 to 3.23, and reduces speedup from $1.93\times$ to $1.68\times$. Removing continuation reuse together with early stopping lowers acceptance length to 3.87 and speedup to $1.84\times$, while removing action noise yields 3.87 and $1.90\times$. Although the latter two variants have the same acceptance length at the reported precision, their speedups differ. On this target, the complete configuration achieves both the highest acceptance length and the highest speedup, indicating that each component contributes when the other two remain enabled.

\subsection{Continuation Reuse on Language Models}
\label{sec:supp_llm_continuation}
\begin{table}[t]
\centering
\caption{\textbf{Continuation reuse on language models.}
We report acceptance length, speedup relative to AR, and accuracy (\%).
+ Reuse adds continuation reuse to DSpark under the same tree budget.}
\label{tab:supp_llm_continuation}
\tabcapbelow
\setlength{\tabcolsep}{7.5pt}
\scriptsize
\resizebox{\linewidth}{!}{%
\begin{tabular}{llrrr}
\toprule
Benchmark & Method & $\tau\uparrow$ & Speedup$\uparrow$ & Acc.$\uparrow$ \\
\midrule
\multicolumn{5}{c}{\textit{Qwen3-4B}} \\
\midrule
\multirow{3}{*}{GSM8K} & AR & --- & $1.00\times$ & 91.89 \\
& DSpark & 7.455 & $4.44\times$ & 91.51 \\
& + Reuse & 7.454 & $4.44\times$ & 91.74 \\
\midrule
\multirow{3}{*}{HumanEval} & AR & --- & $1.00\times$ & 83.54 \\
& DSpark & 7.089 & $4.28\times$ & 85.37 \\
& + Reuse & 7.046 & $4.25\times$ & 83.54 \\
\midrule
\multicolumn{5}{c}{\textit{Gemma-4-12B-it}} \\
\midrule
\multirow{3}{*}{GSM8K} & AR & --- & $1.00\times$ & 84.31 \\
& DSpark & 7.406 & $5.42\times$ & 84.61 \\
& + Reuse & 7.393 & $5.41\times$ & 84.31 \\
\midrule
\multirow{3}{*}{HumanEval} & AR & --- & $1.00\times$ & 95.73 \\
& DSpark & 7.140 & $5.19\times$ & 95.73 \\
& + Reuse & 7.103 & $5.16\times$ & 96.34 \\
\bottomrule
\end{tabular}%
}
\tabbelow
\end{table}

We evaluate Qwen3-4B~\cite{yang2025qwen3} and Gemma-4-12B-it~\cite{team2026gemma} on all 1,319 GSM8K~\cite{cobbe2021training} problems and 164 HumanEval~\cite{chen2021evaluating} tasks. We use publicly released five-layer DSpark draft checkpoints\footnote{\url{https://huggingface.co/deepseek-ai/dspark_qwen3_4b_block7}}\footnote{\url{https://huggingface.co/deepseek-ai/dspark_gemma4_12b_block7}} without retraining, extending drafting to trees with 60 nodes. Decoding is greedy in BF16.

\cref{tab:supp_llm_continuation} shows no speedup gain from reuse in these settings: acceptance length changes by at most 0.043 tokens, and speedup is unchanged or slightly lower. Small accuracy differences are consistent with numerical effects of BF16 tree verification, as discussed in \cref{sec:supp_precision}.

DSpark~\cite{dspark} explains how preceding tokens constrain subsequent predictions. In language generation, these dependencies can reflect grammatical and structural constraints: changing a preceding token may alter which continuations remain appropriate. Correcting the prefix may therefore make continuations from rejected branches less useful, offering a possible explanation for the limited gain here.
These results suggest that continuation reuse benefits from the structure of action sequences in VLAs, where changing a preceding token can leave subsequent predictions useful.

\subsection{Relaxed Verification on OpenVLA}
\label{sec:supp_openvla_comparison}
\begin{table}[t]
\centering
\caption{\textbf{OpenVLA on LIBERO Goal.} Strict uses exact verification; Relaxed uses the acceptance rules described in the text. SR is in percent. A dash denotes an unreported result.}
\label{tab:supp_openvla_comparison}
\tabcapbelow
\setlength{\tabcolsep}{8pt}
\scriptsize
\resizebox{\linewidth}{!}{%
\begin{tabular}{llrrr}
\toprule
Method & Verification & $\tau\uparrow$ & Speedup $\uparrow$ & SR \\
\midrule
AR & - & 1.00 & $1.00\times$  & 76.2 \\
\midrule
SpecVLA & Strict & 1.70 & $0.99\times$  & 77.6 \\
SpecVLA & Relaxed & 2.38 & $1.23\times$  & 75.0 \\
KERV & Relaxed & 2.36 & $0.96\times$  & 73.2 \\
\midrule
DSpark & Strict & 1.80 & $1.10\times$  & 74.2 \\
DSpark & Relaxed & 3.19 & $1.59\times$  & 76.8 \\
\midrule
\rowcolor{aliceblue}
VSpark & Strict & 2.26 & $1.27\times$  & 76.8 \\
\rowcolor{aliceblue}
VSpark & Relaxed & \textbf{3.61} & $\mathbf{1.67}\times$ & 75.2 \\
\bottomrule
\end{tabular}%
}
\tabbelow
\end{table}

Unlike the frequency~\cite{fast} and learned vector-quantized~\cite{actioncodec,g05} action tokenizers in our main experiments (\cref{sec:experiments}), OpenVLA~\cite{openvla} quantizes each action dimension into ordered scalar bins. This gives bin distance a direct interpretation in action space, enabling relaxed verification with a distance threshold. We compare SpecVLA~\cite{specvla}, KERV~\cite{kerv}, DSpark, and VSpark on LIBERO Goal over 500 episodes per setting.\footnote{We exclude HeiSD~\cite{heisd} because no public implementation was available at the time of our experiments.} We use the public SpecVLA\footnote{\url{https://github.com/PineTreeWss/SpecVLA}} and KERV\footnote{\url{https://github.com/zhengzihaoPKU/KERV}} implementations. 
Both SpecVLA and KERV use the EAGLE-2~\cite{eagle2} draft architecture.
For SpecVLA, we use its publicly released draft checkpoint. KERV follows the same draft training approach as SpecVLA, with matching draft architecture and optimizer settings in their public implementations. Since a KERV draft checkpoint is not explicitly provided, we use the public SpecVLA checkpoint with KERV's implementation.

\subpara{Evaluation protocol.}
OpenVLA generates seven tokens per action. The public SpecVLA implementation checks the stopping condition only after committing a full accepted block, allowing tokens beyond this boundary to be counted and inflating the reported acceptance length. We correct this behavior by limiting generation and acceptance-length accounting to the seven action tokens. Before the correction, our reproduction yielded acceptance lengths of 2.03 under strict verification and 2.95 under relaxed verification, close to the results reported by SpecVLA. After the correction, these decrease to 1.70 and 2.38, respectively, as reported in \cref{tab:supp_openvla_comparison}. Strict denotes exact verification. For SpecVLA, DSpark, and VSpark, Relaxed follows SpecVLA's LIBERO Goal setting, allowing draft and target action values to differ by at most 9 quantization bins. For KERV, we follow its paper's setting and adapt the relaxation threshold from 15 to 5. These relaxed settings are separate from lossless decoding.

\subpara{Results.}
\cref{tab:supp_openvla_comparison} shows that VSpark reaches $1.27\times$ speedup with exact verification, compared with $1.10\times$ for DSpark and $0.99\times$ for SpecVLA. Under Relaxed, VSpark reaches $\tau=3.61$ and $1.67\times$ speedup, compared with 3.19 and $1.59\times$ for DSpark and 2.38 and $1.23\times$ for SpecVLA. Using KERV's public implementation, we obtain a higher acceptance length ($\tau=2.36$), but its $0.96\times$ speedup remains below AR in our experiments.

\stopcontents[supp]

\clearpage

{
    \small
    \bibliographystyle{ieeenat_fullname}
    \bibliography{shortstrings,main}
}

\else

\fi
\end{document}